\RequirePackage{fix-cm}
\documentclass{svjour3}                     % onecolumn (standard format)
\smartqed  % flush right qed marks, e.g. at end of proof
\usepackage{array}
\usepackage{graphicx}
\usepackage{hyperref}
\usepackage[export]{adjustbox}
\usepackage{enumitem}
\usepackage{makecell}
\setlist{nosep}
\usepackage{url} % support \url{} command
\usepackage{amsmath} % math extensions
\usepackage{amsfonts} %...and ams math fonts
\usepackage{multicol} % Allow spanning cells in tables
\usepackage{multirow}
\usepackage{graphics} % .jpg, .png, .pdf image import 
\usepackage[table]{xcolor}

\newcommand{\bi}{\begin{itemize}[leftmargin=*]}
\newcommand{\ei}{\end{itemize}}
\newcommand{\be}{\begin{enumerate}[leftmargin=*]}
\newcommand{\ee}{\end{enumerate}}

\usepackage{mdframed}
\newcounter{rules}
\newmdenv[%
    linewidth=0.6pt,
    linecolor=black,
    outerlinewidth=0pt,
       firstextra=false, nobreak=true,
    settings={\global\refstepcounter{rules}},
]{myrules}

\newcommand{\rules}[1]{
    \vspace{-0.5em}
    \begin{myrules}
	#1
    \end{myrules}
}

\usepackage[skins]{tcolorbox}
\usepackage{subcaption}
\usepackage{multirow}

\newcommand{\goalStatement}{to streamline story point estimation by evaluating a comparative learning-based framework for calibrating project-specific story point prediction models}

\begin{document}

\title{Efficient Story Point Estimation With Comparative Learning}


\titlerunning{Code Search with Dual Encoders}        % if too long for running head

\author{Monoshiz Mahbub Khan \and
        Xiaoyin Xi \and
        Andrew Meneely \and
        Yiming Tang \and
        Zhe Yu
}


\institute{M. Khan, X. Xi, A. Meneely, Y. Tang, Z. Yu\at
              Rochester Institute of Technology\\
              Rochester, New York\\
              \email{mk7989,xx4455,axmvse,yxtvse,zxyvse@rit.edu}
}


\maketitle

\begin{abstract}
Story point estimation is an essential component of agile software development. Story points are unitless, project-specific effort estimates that help developers plan their sprints. Traditionally, developers have collaboratively estimated story points using planning poker or other manual techniques. While the initial calibration of the estimates for each project is helpful, once a team has converged on a set of precedents, story point estimation can become tedious and labor-intensive. 
Machine learning can reduce this burden, but only with sufficient context from the historical decisions made by the project team. That is, state-of-the-art models, such as GPT2SP and FastText-SVM, only make accurate (within-project) predictions when they are trained on data from the same project.
The goal of this study is \textit{\goalStatement}. 
Instead of assigning a specific story point value to every backlog item, developers are presented with pairs of items and asked to indicate which item requires more effort. Using these comparative judgments, a machine learning model was trained to predict the story point estimates. We empirically evaluated our technique using data from 23,313 manual estimates across 16 projects. The model trained on comparative judgments achieved, on average, a 0.34 Spearman's rank correlation coefficient between its predictions and the ground truth story points. This is similar to, if not better than, the performance of a state-of-the-art regression model trained on ground truth story points. Through human subject experiments, the advantages of comparative judgments were validated: higher confidence, lower annotation time, and comparable agreement were observed for comparative judgments compared to direct ratings. In summary, the proposed comparative learning approach is more efficient than regression-based approaches, given its better performance, lower required annotation time, and higher training data reliability. 
\end{abstract}


% \vspace{-6ex}
\section{Introduction}
% Trying out something new here - still a wIP. Need to vary how the sentences start for one... but I need to step away
Accurate and efficient effort estimation is essential in agile software development. Story points are unitless, project-specific estimates that help developers communicate their assumptions, discuss effort costs, and plan their upcoming sprints \cite{pasuksmit2024systematic,tawosi2022investigating}. The process of assigning story points is intended to be a collaborative activity for teammates, such as in Planning Poker by Grenning~\cite{grenning2002planning}. As one of the most widespread story point assignment practices, Planning Poker involves the team assigning a story point to each backlog item and then debating any disagreements until an agreement is reached. The benefit of this activity, and others like it \cite{williams_protection_2009,williams_protection_2010}, is to uncover assumptions and misunderstandings and promote knowledge transfer. Later, researchers proposed complex formulas and algorithms to estimate these story points \cite{abrahamsson2007effort,bhalerao2009incorporating}. When used as a proxy for complexity, difficulty, and effort, story points are useful for prioritizing backlog items~\cite{pasuksmit2024systematic}.

However, once the initial benefits are met, story point estimation becomes tedious and labor-intensive. Researchers and practitioners suggest using past precedents to speed up the discussion process~\cite{cohn2005agile,grenning2002planning,williams_protection_2009}, as well as using Fibonacci-like numbers to limit the options~\cite{grenning2002planning,williams_protection_2009,williams_protection_2010}. If the team lacks knowledge of these precedents, the process will slow down~\cite{fu2022gpt2sp}.

State-of-the-art approaches \cite{choetkiertikul2018deep,fu2022gpt2sp} address this problem by training machine learning models to predict story points. However, because the story points estimated by humans are highly subjective and contextual, such machine learning models perform poorly in cross-project predictions~\cite{choetkiertikul2018deep,fu2022gpt2sp} because these human decisions from individual projects are not transferable to other or new ones. Consequently, prediction only works within a project, and human decisions are still required as training data for each new software project.

The goal of this study is \textit{\goalStatement}. We propose replacing the direct estimates of story points with \textbf{comparative judgments}. Instead of manually estimating a story point for each backlog item, developers consider a pair of items and decide which one should have a higher story point. According to the law of comparative judgments~\cite{thurstone1927law}, providing such comparative judgments can be much less strenuous than assigning specific values. In fact, such benefits have been observed for many other decision-making processes in previous research on comparative judgments (see Section~\ref{sec:2.3}). \\

% \yiming{Replace the following sentences with the summary of RQs}

% We developed a novel comparative learning framework to model these comparative judgments. On 16 software projects, we compared the performance of the proposed approach with two state-of-the-art regression models, a baseline regression model using the same encoder architecture as our comparative learning model, and a state-of-the-art linear support vector machine (SVM)-based model also trained on comparative judgments. The results show that our comparative learning approach outperforms all state-of-the-art approaches and achieves performance similar to, if not better than, its regression counterpart.

% We conducted similar experiments in cross-project settings and observed similar results: the comparative learning model performed as well as or better than the state-of-the-art regression models, while the overall performance decreased compared to within-project settings.

% In addition, we conducted active learning experiments to explore whether the proposed models can show comparable performance with less data to further reduce the strain on human judges to collect labeled data.

% Finally, we conducted small-scale human subject experiments to explore how the findings from the machine learning experiments carry over to practical settings. Our results validate the advantages of comparative judgments: they require less annotation time and are more consistent with the ground truth than direct estimates.

With the development of a novel comparative learning framework to model comparative judgments, this work explores on the following research questions:\\

\begin{itemize}
    \item \textbf{RQ1 Is our proposed SBERT-Regression model more capable than state-of-the-art models when predicting the relative order of the items in terms of their story points?}
    As described in Section~\ref{sec:encoder}, we used a new regression model, SBERT-Regression, based on pre-trained SBERT embeddings to predict story points. This research question explores whether this regression model outperforms the latest regression models in story point estimation by comparing its performance to GPT2SP \cite{fu2022gpt2sp} and FastText-SVM \cite{atoum2024enhancing}. Figure~\ref{fig:diff} illustrates the architectures of the three regression models.\\
    
    \item \textbf{RQ2 Can models trained on comparative judgments perform similarly to, if not better than, models trained on direct ratings?}
    This research question explores whether models trained on comparative judgments can achieve a prediction performance similar to that of models learned from direct ratings. To do this, we compared the best-performing SBERT-regression model with (1) LinearSVM-comparative: a baseline comparative model from Qian et al.~\cite{qian2015learning}; and (2) SBERT-comparative: the proposed comparative learning model using the same SBERT-regression encoder architecture but optimizing a hinge loss based on the comparative judgments. To ensure that the comparative models required less human effort, we randomly generated the same number of comparative judgments as the training data points of the regression model.\\
    
    \item \textbf{RQ3 Do more training data pairs lead to better performance? }
    In RQ3, we experimented with increasing the number of training pairs ($k\in[1,10]$ times the training data points of the regression model) to determine whether the prediction performance could be improved with a larger annotation budget. \\
    
    \item \textbf{RQ4 How does the proposed model perform in cross-project predictions?}
    In this RQ, whether comparative judgment can have a positive impact on cross-project predictions, and how these results differ from within-project predictions.\\
    
    \item \textbf{RQ5 Is it faster and easier for human developers to provide the comparative judgments than direct story point estimates?}
    This RQ explores whether it is easier and faster for human judges to make comparative judgments than to make direct story point estimates through human subject experiments.\\
    
    \item \textbf{RQ6 For story point estimation, are comparative judgments more reliable than direct story point estimates?}
    This RQ compares the reliability of comparative judgments with direct estimates in the human subject experiments.\\
\end{itemize}

% Thus, it is promising to streamline story point estimation using comparative learning. The contributions of this study are as follows:
The contributions of this study are as follows:

\begin{itemize}[noitemsep,topsep=0pt]
    
    \item A new regression model with pre-trained SBERT embeddings outperforming state-of-the-art approaches in the estimation of story points, in both within-project (RQ1) and cross-project (RQ4) settings, showing an average improvement of 22.1\% and 17.4\% in terms of Spearman's rank correlation coefficient over state-of-the-art models.\\
    
    \item A novel comparative learning framework to model comparative judgments that performed similarly to, if not better than, the regression model with the same SBERT-based architecture, in both within-project (RQ2) and cross-project (RQ4) settings. These improvements were observed to be approximately 8.1\% for within-project experiments in terms of Spearman's rank correlation coefficient over state-of-the-art models, while cross-project experiments showed comparable results.\\

    \item Showcasing how the amount of data influences model performance for modeling comparative judgments (RQ3).\\
    
    \item Conduct human-subject experiments to show how these findings translate to real-life scenarios. The experiments showed that comparative judgments, on average, took 14.69\% less time while having 8.5\% more confidence from the participants (RQ5). The experiments also showed that direct and comparative judgments showed comparable inter-rater agreement (RQ6) on average across participants and groups.\\
    
    \item Code and data used in this study are publicly available \footnote{ \url{https://github.com/hil-se/EfficientSPEComparativeLearning}}.\\
    
\end{itemize}

The rest of this paper is organized as follows. Section~\ref{sec:related_works} discusses relevant works in story point estimation and comparative judgments. Followed by Section~\ref{sec:methods} presenting the methodology behind our proposed model structure and comparative framework. Section~\ref{sect:within} answers RQ1, RQ2, and RQ3 using within-project simulations on 16 open source software projects while Section~\ref{sect:cross} answers RQ4 with simulations in a cross-project setting. Section~\ref{sec:human_exp} describes two human subject experiments to validate the advantages of comparative judgments and answer RQ5 and RQ6. Finally, we discuss the potential threats to the validity of this study in Section~\ref{sec:threats} before arriving at our conclusions in Section~\ref{sec:future_work_and_conclusion}.

\section{Related Works}
\label{sec:related_works}
The following subsections explore traditional and manual story point calculation methods, previous work on estimating story points using machine learning methods and previous work on comparative judgment across different domains.

\subsection{Manual story point estimation}

% \yiming{Analogous Estimation is another solution. }

Manual agile story point estimation methods involve considering the information at hand about the items and previously calculated story points for similar items to approximate a new item's story point. Cohn \cite{cohn2005agile} investigated some of these conventional methods. One way to approximate these story points is to consider the number of lines of code or function points in the item or required to complete the item. Integrating information about similar story points has also been a significant process in other methods, such as estimation by analogy, triangulation, and disaggregation.\\

\bi
\item
Estimating by analogy involves comparing the item at hand with a past item with a known story point. The intuition is that if the items are similar, their story points should also be similar. \cite{chauhan2021story} \cite{shepperd1997estimating}\\
\item
Triangulation goes one step further and compares the new item with multiple past items in an attempt to triangulate into a more precise estimation.\cite{cohn2005agile}\\
\item
Disaggregation is another method that is primarily used for complex items. It involves breaking down a new item into smaller items \cite{coelho2012effort}, estimating their story points using estimation by analogy or triangulation, and then adding them up. One disadvantage of these methods is that they only consider items, but not the context or the humans who would be working on them. \\
\ei

Planning poker is a popular story point estimation method that solves this shortcoming. Planning poker \cite{grenning2002planning} is an item story point estimation method in which a team of people involved in tackling or solving these items gathers and assigns story points to each item individually. After revealing the points assigned by each team member, it was observed whether everyone assigned the same score. In the case of an agreement, that score would be the final score. However, in the case of a disagreement where different team members might have assigned different scores, the item, its possible solutions, the required and available resources, and other contextual information are used for debate until a consensus is reached for the final story point for that item. Planning poker has the advantage of considering contextual information about resources and human efforts, not just information about the item at hand. However, it is not without its drawbacks. Planning poker can take a significant amount of time and resources to thoroughly go through all the items and reach agreements for all of them. Moreover, the points assigned to a task can vary depending on the knowledge and experience of human judges \cite{mahnivc2012using}, resulting in inconsistencies in their decisions.

% There have been efforts to streamline this process, and reduce the amount of human involvement and resources spent in estimating these story points through various automated algorithms and complex formulae \cite{abrahamsson2007effort,bhalerao2009incorporating}. Certain modern practices involve using historical story point data collected through these past methods to train complex machine learning-based models to make these predictions. However, the most commonly used method in practice remains planning poker.

\textit{In summary, traditional manual methods of story point estimations require a significant amount of time and effort from human developers. These methods often result in inconsistent estimations owing to the variance in each project's context and each developer's experience. Despite all this, planned poker is still widely used because of its accuracy in consistent contexts.}

\subsection{Predicting story points with machine learning}
% ML/DL based methods.
Contemporary methods utilize a variety of machine learning or deep learning-based methods and make use of historical story point data to make predictions with reasonable accuracy and lower strain on resources and time. 
GPT2SP \cite{fu2022gpt2sp} and FastText-SVM \cite{atoum2024enhancing} are considered state-of-the-art models for this task, as they show much higher performance than previous works. Therefore, these two models were used as the baselines in our experiments.

Abrahamsson et al. \cite{abrahamsson2011predicting} proposed several regression models on two industrial Agile software projects for story point estimation. The data used in this study were formatted to include features such as the presence of specific words and item priority. Porru et al. \cite{porru2016estimating} treated this as a classification task. This study integrated Term Frequency-Inverse Document Frequency (TF-IDF) in the training process and concluded that a larger number of training instances is required for a model to make consistent and accurate predictions. In contrast, Scott and Pfahl \cite{scott2018using} integrated information on developers making decisions into their dataset. They observed that their Support Vector Machine (SVM) model performed better with this type of information present in their dataset compared to using only text data. On the other hand, Soares \cite{soares2018effort} used an encoder-decoder architecture-based text classification model to estimate story points with reasonable success.

Tawosi et al. \cite{tawosi2022investigating} used a Latent Dirichlet allocation (LDA) model to generate word embeddings for the issue text and trained a k-means clustering model. Three different methods were used to estimate the story point for the testing point: (i) through the mean of the story points in the predicted cluster, (ii) through the median of the story points in that cluster, and (iii) through the story point of the closest point in that cluster. Yal{\c{c}}{\i}ner et al. \cite{yalcciner2024enhancing} used this dataset from Tawosi et al. \cite{tawosi2022investigating}, and builds a siamese BERT like structure, SBERT for feature extraction and combines it with various gradient-boosted tree algorithms and contrastive learning to show performance similar to various baseline methods.

Choetkiertikul et al. \cite{choetkiertikul2018deep} introduced a dataset for story point estimation across different projects, and trains their model Deep-SE on this data for story point estimation. This model was trained on each project's texts individually and used the model's generated word embeddings with the labels from the data, treating it as a regression task. Training the model on each project's text data at a time results in accurate predictions for within-project experiments but shows poorer performance for cross-project experiments. The data used by Choetkiertikul et al. \cite{choetkiertikul2018deep} were used in our experiments.

Fu and Tantithamthavorn \cite{fu2022gpt2sp} used their model GPT2SP on the same data as Choetkiertikul et al. \cite{choetkiertikul2018deep}. This study used a GPT2 based language model that uses a byte-pair encoding subword tokenization method to split rare words into subword units, effectively reducing the vocabulary size significantly. With a more compact vocabulary, the features used in their model were less sparse and more meaningful. This work built a GPT2 based structure and made use of a masked multi-head self-attention mechanism to train on this data and showed much better performance than previous works in within-projects experiments, but still showed higher error for cross-project experiments. GPT2SP is considered a state-of-the-art model for story point estimation.

Li et al. \cite{li2024fine} collected and made use of complementary numerical data to make more accurate predictions. This study involved interviewing enterprise employees to collect resource-related numerical features on the issues it trained and tested on. The model showed better performance than state-of-the-art models for the other projects but fell short for publicly available data. Moreover, the requirement for additional numerical data made this work not generalizable without similar additional data for any new project, which would require more human effort.

Atoum et al. \cite{atoum2024enhancing} made use of pre-trained word embeddings and simple regression models to show impressive performance on small datasets. This study evaluated various pretrained word embedding and regression models to identify the best performing pair. Their final model used a pre-trained FastText word embedding model \cite{fasttext} and an SVM model for the regression. This study showed a performance similar to that of state-of-the-art models. However, the use of FastText word embeddings makes the text-processing step of this method more resource-intensive than that of other methods.

\textit{In summary, GPT2SP \cite{fu2022gpt2sp} and Atoum et al. \cite{atoum2024enhancing} showed impressive performance for story point estimation using machine learning methods. However, machine learning-based story point estimation models require a large amount of labeled data from the same project to achieve reasonable performance. These methods tend to perform poorly when the training data are not obtained from the same project. In other words, to achieve usable performance, machine learning-based methods require sufficient amount of human estimates from the same project.}

\subsection{Comparative judgment}\label{sec:2.3}
Comparative judgment has a long history of research across various fields and domains. In the education domain, comparative assessments of students have been shown to be more reliable \cite{bramley2015investigating} than individual assessments. Additionally, according to Verhavert et al. \cite{verhavert2019meta}, ``comparative judgments are considered to be easier and more intuitive, as people generally base their decisions on comparisons, either consciously or unconsciously." F{\"u}rnkranz and H{\"u}llermeier \cite{furnkranz2010preference} also mentioned that such comparative judgments are more intuitively appealing to human judges, claiming that it is often easier to compare alternatives pairwise. An easier and more intuitive task also implies lower effort and cognitive burden on human judges, according to the law of comparative judgment~\cite{thurstone1927law}.

Modeling comparative judgments with machine learning models is challenging because the dependent variable is always associated with two data objects, indicating their relative relationships. Thus, traditional classification or regression algorithms cannot be directly applied to model comparative judgments.

F{\"u}rnkranz and H{\"u}llermeier \cite{furnkranz2003pairwise} is one of the earlier works in modeling such pairwise comparison, using a decision tree to rank rational agents' actions in uncertain environments. Brinker (2004) \cite{brinker2004active} approached a ranking problem as a binary classification problem, and used active learning methods on synthetic data to show reasonably high performance. F{\"u}rnkranz and H{\"u}llermeier \cite{furnkranz2010preference} brought up the idea of approaching order classification or ranking tasks in terms of label preferences. Indeed, using the labels of each data point to define their pairwise order and using that information to learn the relative order of all data points is a concept that heavily inspires our work here.

Numerous studies have adopted a pairwise approach from various perspectives. Burges et al. \cite{burges2005learning} adapted the pairwise data format to build a recommendation system. The data points were trained in pairs. During testing, for each query, all possible pairs were formed, and their model output scores were generated and sorted to find the desired results from the pool of data points. On the other hand, F{\"u}rnkranz and H{\"u}llermeier \cite{furnkranz2010preference} examined a few different methods of classification tasks by reformatting the target outputs into pairs. For example, for labels ``A," ``B" and ``C," it might reformat them into pairs of ``A" and ``not A," ``B" and ``not B," and ``C" and ``not C," and conduct binary classification on each pair of these labels at a time.

Qian et al. \cite{qian2015learning} reformatted regression data into pairwise data and focused mainly on data with numerical attributes. This study transformed data with separate data points and continuous target values into pairwise data in the form of $x$, where $x = A - B$, $A$ and $B$ are different data points, and the target label is $y=1$ if $A>B$ and $y=-1$ if $A<B$. An SVM was used on this data to make predictions on new data. This approach showed promising results for data with numerical attributes, but showed much poorer performance for other types of data, such as text word embeddings, where the individual values have no significance on their own, and as such, $A-B$ is not as meaningful as it would be for numerical data. This study is the only previous study that takes a similar comparative judgment-based approach to ours. Therefore, this study was used as a baseline model for comparison with our model.\\

\textit{In summary, there have been multiple works throughout the years to learn a model from pairwise inputs or comparative judgments. However, most of these studies rely on a query or an anchor with a pair of items to build a triplet; in information retrieval, the oracle is usually one of the items with a better retrieval result in terms of the query (anchor) than the other items. To the best of our knowledge, only Qian et al. \cite{qian2015learning} solved a similar problem of comparing two items without a query or an anchor. However, this work limits its learner to a linear SVM, whereas our proposed comparative learning framework can utilize any regression structure, including deep neural networks. This gives us more freedom and capability to tackle complicated problems.}

\section{Methodology}
\label{sec:methods}
Traditionally, story points are estimated through processes such as planning poker, where developers discuss and decide the exact number of story points for each item in the backlog. Here, in the proposed comparative judgment-based process, developers would instead examine a pair of items at a time and decide which one requires more effort (thus a higher story point). Figure~\ref{fig:comp_judg} summarizes this process. For an initial backlog of $n$ items, the process is as follows:

\begin{figure*}[h!]
    \centering
    % \hspace{-2cm}
    \includegraphics[width=0.75\linewidth]{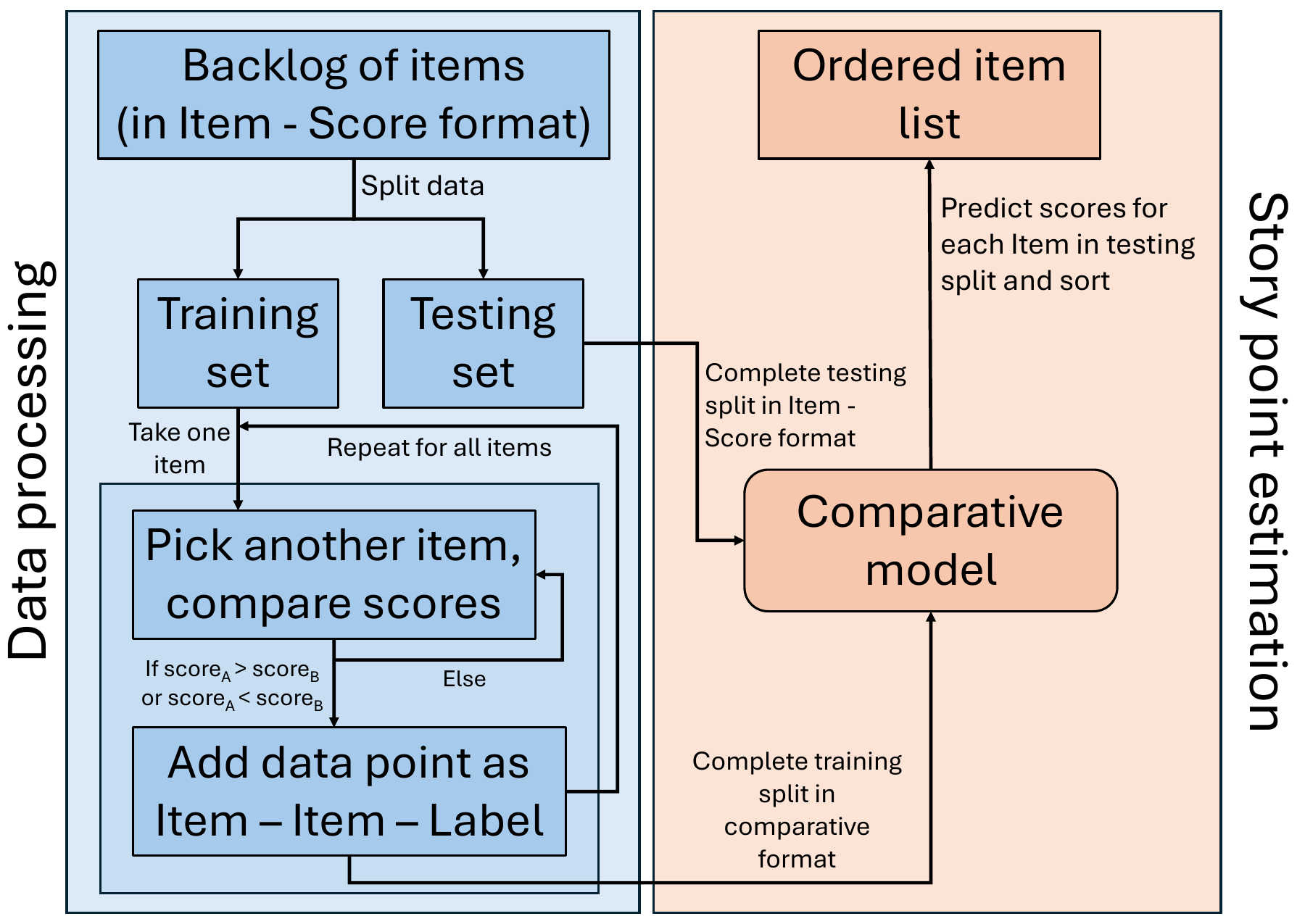}
    \caption{Story point estimation with comparative judgments}
    \label{fig:comp_judg}
\end{figure*}

\begin{enumerate}[leftmargin=10mm]
    \item[Step 1] For each item in the backlog, randomly pick $k$ other items without replacement from the backlog to form $k$ pairs of items ($item_A, item_B$). We ensured that there were no duplicate pairs in this step.
    \item[Step 2] Developers provide a comparative judgment $y$ on each pair. The numerical mapping of the judgment is shown in \eqref{eq:y}.

\begin{equation}\label{eq:y}
        y =  \left\{
        \begin{aligned}
         1, & \quad\text{when }  item_A \text{ requires more effort than }item_B\\
         -1, & \quad\text{when }  item_B \text{ requires more effort than }item_A\\
        \end{aligned}\right. \\
\end{equation}

    \item[Step 3] At the end of this process, there are $k\cdot n$ pairs of backlog items with a comparative judgment label $y$ attached. Every item appears in a pair at least $k$ times, and there are no duplicate or mirror pairs. For example, if ($item_A, item_B$) is chosen as a pair, ($item_B, item_A$) is not chosen.
    \item[Step 4] These comparative judgment data ($item_A,item_B,y$) are used to train a machine learning model $f_{\theta}(x)$ with \textbf{comparative learning} described in Section~\ref{sec:framework}.
    \item[Step 5] The model predicts a score $f_{\theta}(x)$ to rank every item in the initial backlog. Developers can use the predicted scores and ranks of each item for sprint planning.
    \item[Step 6] In future sprints, new backlog items can be directly assigned with scores and ranks by the model without human intervention. \\
\end{enumerate}

In summary, for practical scenarios, developers would make $k\cdot n$ comparative judgments instead of $n$ story point estimates. Then, a model is trained on the developers' comparative judgments and sorts the individual items into a ranked order. Because the learned model has no information on the actual story point values, the predicted scores can be quite different from the story points the developers might assign. However, because story points are unitless and the primary goal is to prioritize the backlog, this approach simplifies the process and saves effort (especially when $k=1$).

\subsection{Encoder structure}\label{sec:encoder}
\begin{figure*}[h!]
    \centering
    \includegraphics[width=0.85\linewidth]{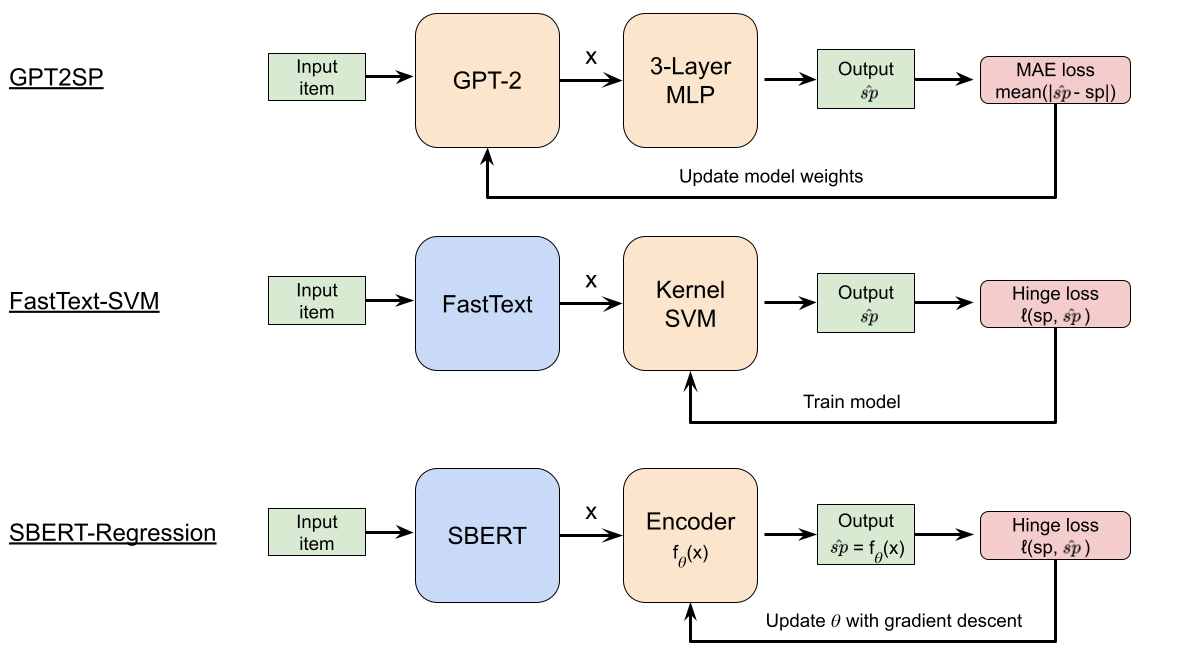}
    \caption{Regression model architectures.}
    \label{fig:diff}
\end{figure*}

The text of the original backlog items is passed through an SBERT \cite{reimers-2019-sentence-bert} model to represent the items using sentence embeddings $x$. These sentence embeddings $x$ are then passed onto a dense layer that maps the sentence embeddings $x$ to the scalar output $f_{\theta}(x)$. This encoder structure is used as (1) our baseline regression model (SBERT-Regression), which outperformed state-of-the-art baselines GPT2SP and FastText-SVM (shown later in Figure~\ref{fig:diff}), and (2) the encoder $f_{\theta}(x)$ in our comparative learning model (SBERT-Comparative), as shown in Figure~\ref{fig:training}.

\subsection{Comparative learning framework}\label{sec:framework}
\begin{figure*}[!t]
    \centering
    % \hspace{1in}
    \includegraphics[width=0.9\linewidth]{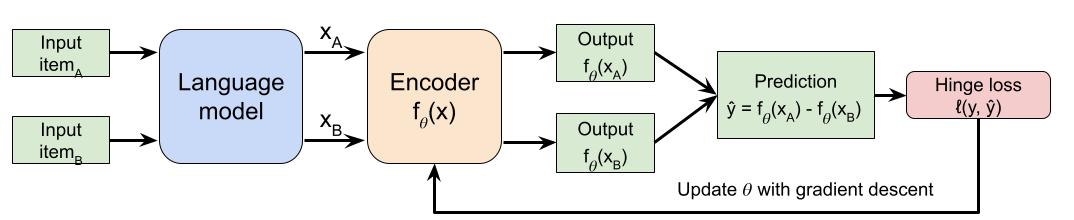}
    \caption{Proposed comparative learning framework}
    \label{fig:training}
\end{figure*}

The training step of our approach is illustrated in Figure~\ref{fig:training}. This is a general framework for modeling comparative judgments, where any regression model can be used as the encoder $f_{\theta}(x)$. For an input training pair with the corresponding comparative judgment label ($item_A$, $item_B$, $y$), the encoder generates an individual output score $f_{\theta}(x)$ for each item.
The prediction of the comparative relationship between $item_A$ and $item_B$ is calculated as:
\begin{equation}\label{eq:yhat}
\hat{y} = f_{\theta}(x_A) - f_{\theta}(x_B).
\end{equation}

Next, the hinge loss shown in \eqref{eq:loss} is used to update the weights $\theta$ via stochastic gradient descent.

\begin{equation}\label{eq:loss}
\ell(y, \hat{y}) = max(0, 1 - (y\cdot\hat{y})).
\end{equation}
Here, the hinge loss is chosen as the loss function because it is always $0$ as long as the predicted relationship is in the correct direction $\text{sgn}(y) = \text{sgn}(\hat{y})$ and $|\hat{y}|\ge 1$. This aligns with the fact that from the human oracles $y$, we only know which item requires more effort, but no information on the exact amount of effort in difference.

\begin{figure*}[h!]
    \centering
    % \hspace{-0.7cm}
    \includegraphics[width=0.85\linewidth]{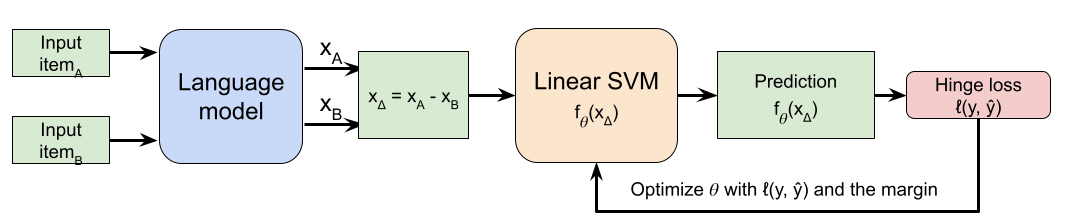}
    \caption{LinearSVM-Comparative from Qian et al.~\cite{qian2015learning} with a pre-trained language model.}
    \label{fig:linearsvm}
\end{figure*}

% In contrast to the pairwise training process, only a single data point is passed to the model during testing. The raw output of the model was considered the final prediction. The scores for all items were then used to sort them.

% \subsection{Active learning}
% \label{method:active}
% Active learning was integrated into the machine learning experiments to explore how the proposed models can show comparable performance even with less data available. Different approaches were adopted for regression and comparable models. In both cases, a portion of the original dataset, $k$, was separated to function as the initial minimum data required to train the model. In our experiments, the number of data points or pairs for the initial state, $k$ was 10. After training the model on this initial amount of data, a certain number of data points or pairs were fed to the model to update its training on each iteration until the pre-decided maximum amount of data was reached. In these experiments, $k=10$ data points or pairs were used to update the model for each iteration.

% The experiments conducted in this work are organized different categories - (i) Machine learning experiments, (ii) Active learning experiments, (iii) Cross-project experiments, and (iv) Human subject experiments.

% \yiming{improve -\textgreater Improve}
%%%%%%%%%%%%%%%%%%%%%%%%%%%%%%%%%%%%%%%%%%%%%%%%%%%%%%%%%%%%%%%%%%%%%%%%%%%%%%%%%%%
\section{Within-project experiments}\label{sect:within}
In light of our proposed comparative judgment-based approach, we examine the following research questions:\\

% \yiming{I think the RQs should be reformulated as questions rather than declarative statements.}
\begin{itemize}
    \item \textbf{RQ1 Is our proposed SBERT-Regression model more capable than state-of-the-art models when predicting the relative order of the items in terms of their story points?}
    As described in Section~\ref{sec:encoder}, we used a new regression model, SBERT-Regression, based on pre-trained SBERT embeddings to predict story points. This research question explores whether this regression model outperforms the latest regression models in story point estimation by comparing its performance with those of GPT2SP \cite{fu2022gpt2sp} and FastText-SVM \cite{atoum2024enhancing}. Figure~\ref{fig:diff} illustrates the architectures of the three different regression models.\\
    \item \textbf{RQ2 Can models learning from comparative judgments perform similarly to, if not better than, models trained on direct ratings?}
    This research question explores whether models learned from comparative judgments could achieve prediction performance similar to that of models learned from direct ratings. To do this, we compare the best performing SBERT-Regression model with (1) LinearSVM-comparative: a baseline comparative model from Qian et al.~\cite{qian2015learning} using the same pre-trained SBERT embeddings. As shown in Figure~\ref{fig:linearsvm}, it first subtracts the embeddings of the two input items and trains a linear SVM model to predict $y$; (2) SBERT-Comparative: the proposed comparative learning model with pre-trained SBERT embeddings shown in Figure~\ref{fig:training}. To ensure that the comparative models require less human effort, we set $k=1$ in this research question so that the size of the training data would be the same for the comparative models and the regression model.\\
    \item \textbf{RQ3 Do more training data pairs lead to better performance? }In \textbf{RQ2}, the same number of comparative judgments and direct story points are used in training ($k=1$) to evaluate whether the comparative learning model is as good as its regression counterpart while requiring less effort in human annotation. In RQ3, we experiment with increasing the number of training pairs $k\in\{1,2,3,4,5,10\}$ to determine whether the prediction performance could be improved with a larger annotation budget. 
\end{itemize}

% \vspace*{-8ex}
\subsection{Data}
This work uses the story point estimation dataset introduced in Choetkiertikul et al. \cite{choetkiertikul2018deep}. Table-\ref{table:data_characteristics} lists the total size and range of story points for each project in this dataset. The dataset consists of backlog item titles and descriptions, as well as their assigned story points, collected through JIRA across 16 projects.

For our experiments, we used the same data splits as Fu et al. \cite{fu2022gpt2sp}. The testing split items and their story points were treated as answer sets during the evaluation. During the pre-processing step, the titles and descriptions were concatenated to form the item texts used for training and testing~\cite{fu2022gpt2sp}.

% \vspace*{-2ex}
\begin{table}[h!]
    \centering
    \caption{Data characteristics}
    \begin{tabular}{|l|r|l|}
        \hline
        Project & Data Size & \makecell{Story point\\ range}\\ \hline
        appceleratorstudio & 2,919 & 1 - 40\\
        aptanastudio & 829 & 1 - 40\\
        bamboo & 521 & 1 - 20\\
        clover & 384 & 1 - 40\\
        datamanagement & 4,667 & 1 - 100\\
        duracloud & 666 & 1 - 16\\
        jirasoftware & 352 & 1 - 20\\
        mesos & 1,680 & 1 - 40\\
        moodle & 1,166 & 1 - 100\\
        mule & 889 & 1 - 21\\
        mulestudio & 732 & 1 - 34\\
        springxd & 3,526 & 1 - 40\\
        talenddataquality & 1,381 & 1 - 40\\
        talendesb & 868 & 1 - 13\\
        titanium & 2,251 & 1 - 34\\
        usergrid & 482 & 1 - 8\\ \hline
    \end{tabular}
    \label{table:data_characteristics}
\end{table}
% \vspace*{-6ex}

\newpage
\subsection{Comparative judgment simulation}
\begin{figure*}[h!]
    \centering
    % \hspace{-2cm}
    \includegraphics[width=\linewidth]{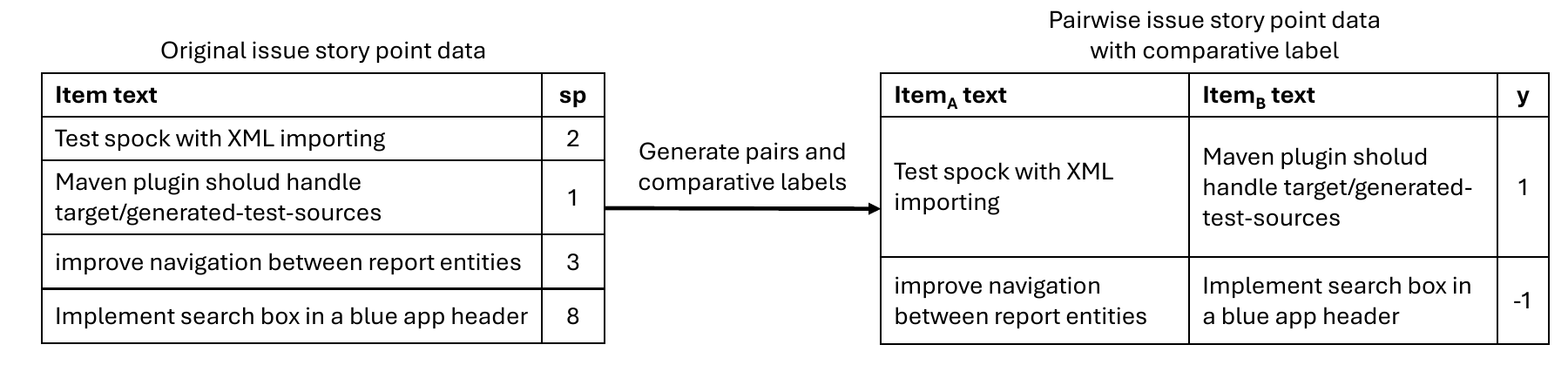}
    \caption{Comparative judgment formatting}
    \label{fig:data_gen}
\end{figure*}

Figure~\ref{fig:data_gen} shows how the data are converted from the original story points to comparative judgments. The original dataset is in the form of $ item - sp $, where sp is the story point for this item. It is converted into comparative judgments of $ item_A - item_B - \hat{y} $, where 
\begin{equation}\label{y_sim}
y =  \left\{
\begin{aligned}
&1, &\text{sp}_A>\text{sp}_B\\
% &0, &\text{when } &x_i \text{ and } x_j \text{ require similar effort}\\
&-1, &\text{sp}_A<\text{sp}_B
\end{aligned}\right..
\end{equation}
For each data point, $k$ random data points from the same split are retrieved, and their story points are compared to form these pairs. Therefore, the number of training data pairs with comparative judgments is $(k \times n)$ when $n$ is the number of training data points with ground truth story points. We used $k=1$ for RQ2 and $k=\{1,2,3,4,5,10\}$ for RQ3.

\subsection{Evaluation metrics}
% Spearman's, Pearson's, MAE, MdAE
Because the story points are unitless and we are evaluating prioritization, Pearson's correlation coefficient $\rho$ and Spearman rank correlation coefficient $r_s$ are utilized to evaluate the predicted order of the data points against the actual order of the data points. We also used Mean Average Error (MAE) scores to compare the performance of the regression models.

\subsubsection{Pearson's correlation coefficient}
For two ordered lists of test data points $f_{\theta}(x)$ and $sp$ with n data points, and means $\overline{f_{\theta}(x)}$ and $\overline{sp}$, respectively, the Pearson's correlation coefficient $\rho$ is calculated as 
\begin{equation}
    \rho={\frac {\sum\limits_{i=1}^{n}(f_{\theta}(x_i)-{\overline {f_{\theta}(x)}})(sp_{i}-{\overline {sp}})}{{\sqrt {\sum\limits_{i=1}^{n}(f_{\theta}(x_i)-{\overline {f_{\theta}(x)}})^{2}}}{\sqrt {\sum\limits_{i=1}^{n}(sp_{i}-{\overline {sp}})^{2}}}}}
\end{equation}
Pearson's coefficient is always within the range of $[-1, 1]$. A positive value implied a positive correlation, whereas a negative value implied a negative correlation. The magnitude of the absolute value of the coefficient indicates the strength of the correlation. The ideal outcome would be a Pearson’s coefficient close to 1 between the encoder outputs and the absolute human ratings.\\

\subsubsection{Spearman's rank correlation coefficient}
The intended use of the framework is to generate a ranked order of the backlog items after training. Therefore, for evaluation, some metrics need to be used that can compare an ordered list against another. Because the emphasis here is on the order of the items on the list, Spearman's rank correlation coefficient was chosen over other metrics such as Pearson's rank coefficient. With $n$ data points, the Spearman's rank correlation coefficient, $r_s$ can be calculated as -
\hspace{-1ex}
\begin{equation}
    r_{s}= {\rho } {\bigl [}R[sp], R[f_{\theta}(x)]\ {\bigr ]},
\end{equation}
where $R[sp]$ and $R[f_{\theta}(x)]$ are the ranks of the ground truth story points $sp$ and the predicted story points $f_{\theta}(x)$, respectively. The value of $r_s$ is in the range $[-1, 1]$, where a value closer to $1$ implies a positive correlation, a value closer to $-1$ implies a negative correlation, and a value closer to $0$ implies no correlation. In other words, a higher value of this coefficient indicates that the model performs well when scoring and ranking the data points.

\subsubsection{Mean Absolute Error}
The Mean Absolute Error (MAE) metric here is the mean of the absolute differences between the predicted and actual story points on the testing data points. A lower MAE score indicates a higher model performance. For $n$ testing data points, where $sp_i$ is the $i$-th actual story point and $f_{\theta}(x_i)$ is the regression model's predicted score for the $i$-th data point, the MAE score can be calculated as
\begin{equation}
    MAE = \frac{\sum\limits_{i=1}^{n} | sp_i - f_{\theta}(x_i) |}{n}
\end{equation}

\subsection{Experimental setup}
% Training/testing process, baselines, hyperparameters etc.
We used the same pre-split training, validation, and testing splits as those used by Fu et al. \cite{fu2022gpt2sp}. All experiments conducted in this study were within-project experiments. Regression experiments on each project were conducted with 20 random repeats, and the mean of each performance metric was reported. Comparative experiments for each $k$ value were conducted with 10 random repetitions, each sampling a different set of data pairs. The detailed parameters for our approaches are listed in Table~\ref{table:hyperparameters}.

\vspace*{-1ex}
\begin{table}[!thp]
    \centering
    \caption{Model hyperparameter values}
    \begin{tabular}{|l|c|c|c|}
        \hline
        & \makecell{SBERT-\\Regression} & \makecell{SBERT-\\Comparative\\(without\\ validation)} & \makecell{SBERT-\\Comparative\\(with\\ validation)}\\ \hline
        
        Train on & \multicolumn{2}{c|}{$\mathcal{D}_{train}\cup  \mathcal{D}_{val}$} & $\mathcal{D}_{train}$ \\\hline
        Validation & \multicolumn{2}{c|}{$\emptyset$} & $\mathcal{D}_{val}$\\\hline
        Loss  & MAE & \multicolumn{2}{c|}{Hinge Loss}\\ \hline
        Max epochs & 600 & 100 & 300\\ \hline
        Learning rate & \multicolumn{3}{c|}{0.001 to 0.000001}\\ \hline
        Optimizer & \multicolumn{3}{c|}{Adam}\\ \hline
        Batch size & \multicolumn{3}{c|}{32}\\ \hline
    \end{tabular}
    \label{table:hyperparameters}
\end{table}

Here, we trained the SBERT-Regression model on the union of the training and validation sets $\mathcal{D}_{train} \cup \mathcal{D}_{val}$ without any validation. This leads to better performance because more information is available during training. Regarding the SBERT-Comparative models, training without validation achieved better performance for lower $k$ values, whereas training with validation achieved better performance for higher $k$ values owing to its ability to prevent overfitting. The LinearSVM-Comparative model was also trained on $\mathcal{D}_{train} \cup \mathcal{D}_{val}$ without any validation because it did not take advantage of a validation set.

We used an Intel i7-4790 3.6 GHz processor, 32 GB of system memory, and an NVIDIA GeForce RTX 2070 with a memory of 8 GB. The hyperparameter values used to train the model are listed in Table-\ref{table:hyperparameters}. The total time required for data formatting and then to train and test on 16 projects is around 4-4.5 hours for all the SBERT-based models, including SBERT-Regression, SBERT-Comparative (with and without validation), and also for LinearSVM-Comparative. This is similar to the 5-hour runtime of GPT2SP.

\subsection{Results}
\subsubsection{RQ1 Comparisons of regression models in story point estimation}
\begin{table*}[!thp]
    \setlength{\tabcolsep}{1pt}
   \footnotesize
    \centering
    \renewcommand{\arraystretch}{1.25}
    \caption{Results for RQ1.}
    \begin{tabular}{|l|c|c|c||c|c|c||c|c|c|}
    \hline
        \multirow{2}{*}{Project} & \multicolumn{3}{c||}{\makecell{FastText-SVM \cite{atoum2024enhancing}}} & \multicolumn{3}{c||}{GPT2SP \cite{fu2022gpt2sp}} & \multicolumn{3}{c|}{\textbf{SBERT-Regression}}\\ \cline{2-10}
        
        & \makecell{$\rho$} & \makecell{$r_s$} & MAE & \makecell{$\rho$} & \makecell{$r_s$} & MAE & \makecell{$\rho$} & \makecell{$r_s$} & MAE\\ \hline

        appceleratorstudio & 0.3019 & 0.2812 & 1.4443 & 0.1665 & 0.1489 & \textbf{1.1288} & \textbf{0.3254} & \textbf{0.3037} & 3.5821\\
        aptanastudio & 0.1808 & 0.1669 & 3.4687 & -0.0442 & 0.0050 & \textbf{2.2506} & \textbf{0.3419} & \textbf{0.2830} & 5.8345\\
        bamboo & 0.1352 & 0.1160 & 0.8425 & 0.0401 & 0.1449 & \textbf{0.6118} & \textbf{0.1768} & \textbf{0.1753} & 0.8148\\
        clover & 0.3223 & 0.2026 & 3.6959 & 0.1477 & 0.2668 & \textbf{2.2477} & \textbf{0.4403} & \textbf{0.4166} & 3.8337\\
        datamanagement & 0.3640 & \textbf{0.4354} & 5.8862 & \textbf{0.5217} & 0.3410 & \textbf{3.8146} & 0.3775 & 0.3909 & 7.0462\\
        duracloud & 0.3252 & 0.3089 & 0.7085 & 0.0459 & 0.2669 & \textbf{0.5697} & \textbf{0.3758} & \textbf{0.4221} & 0.8064\\
        jirasoftware & 0.2576 & 0.1599 & 1.7541 & 0.0433 & 0.1213 & \textbf{1.3050} & \textbf{0.5324} & \textbf{0.4987} & 2.3916\\
        mesos & 0.3182 & 0.3141 & 1.1013 & 0.2651 & 0.3227 & \textbf{0.8894} & \textbf{0.3960} & \textbf{0.3916} & 1.5414\\
        moodle & 0.1893 & 0.1712 & 7.2921 & 0.2799 & \textbf{0.4291} & \textbf{5.3546} & \textbf{0.3499} & 0.3552 & 5.9245\\
        mule & 0.2266 & \textbf{0.2536} & 2.4483 & 0.1593 & 0.1785 & \textbf{1.7313} & \textbf{0.2342} & 0.2459 & 3.3713\\
        mulestudio & \textbf{0.2076} & \textbf{0.1856} & 3.6092 & -0.1373 & -0.1150 & \textbf{1.5632} & 0.1096 & 0.0612 & 5.6180\\
        springxd & 0.3920 & \textbf{0.4231} & 1.6833 & 0.2146 & 0.2261 & \textbf{1.2182} & \textbf{0.3982} & 0.3911 & 2.0790\\
        talenddataquality & 0.2370 & 0.2281 & 3.3422 & \textbf{0.3504} & \textbf{0.4097} & \textbf{1.7789} & 0.2892 & 0.2985 & 2.3418\\
        talendesb & 0.4419 & \textbf{0.4641} & 0.8134 & \textbf{0.4820} & 0.4161 & \textbf{0.4507} & 0.3453 & 0.3550 & 0.9979\\
        titanium & 0.1086 & 0.1059 & 2.2120 & 0.1297 & 0.1582 & \textbf{2.0910} & \textbf{0.1861} & \textbf{0.2264} & 3.8560\\
        usergrid & 0.2265 & 0.2891 & 1.1840 & \textbf{0.3049} & \textbf{0.3851} & \textbf{0.4911} & 0.2016 & 0.1981 & 1.6888\\ \hline
        Average & 0.2647 & 0.2566 & 2.5929 & 0.1856 & 0.2316 & \textbf{1.7185} & \textbf{0.3175} & \textbf{0.3133} & 3.2330\\ \hline
        
    \end{tabular}
    \label{table:issue_baselines}
\end{table*}
Table \ref{table:issue_baselines} compares the performance of our regression approach with those of the GPT2SP and FastText-SVM models on the same story point estimation data. From Table \ref{table:issue_baselines} we observe:
\bi
\item The calculated MAE scores for GPT2SP and FastText-SVM are similar to the scores reported in their works \cite{fu2022gpt2sp} \cite{atoum2024enhancing}. This validates the correctness of our baseline reproduction.
\item SBERT-Regression shows better ranking performance compared to the FastText-SVM and GPT2SP models. Our approach outperforms these baselines on average and for 11 of the 16 projects in terms of either $r_s$ or $\rho$ scores or both. For the remaining five projects, the performance of our approach is similar or marginally lower.
\item SBERT-Regression shows worse performance than the baselines in terms of MAE scores. However, since story points are unitless, the order of the items' story points is far more important than the direct scores; thus, we value the $r_s$ or $\rho$ scores more here.\\
\ei

% \yiming{This RQ seems very short. In general, research questions are not typically framed as concisely. If you intend to keep it this way, you may need to clarify earlier in the paper that the study addresses six RQs, so that the brevity of each individual RQ seems reasonable. In addition, the discussion could be expanded further.}

\rules{\noindent\textbf{Answer to RQ1: }In summary, our approach SBERT-Regression is more capable than state-of-the-art models when predicting the relative order of the items in terms of their story points.}

\subsubsection{RQ2 Comparisons between comparative and regression models}
\vspace*{-3ex}
\begin{table*}[h!]
    \setlength{\tabcolsep}{2.5pt}
   \small
    \centering
    \renewcommand{\arraystretch}{1.25}
    \caption{Results for RQ2.}
    \begin{tabular}{|l||c|c||c|c||c|c||c|c|}
    \hline
        \multirow{2}{*}{Project} & \multicolumn{2}{c||}{SBERT-Reg.} & \multicolumn{2}{c||}{\makecell{LinearSVM-\\ Comp. \cite{qian2015learning}}} & \multicolumn{2}{c||}{\makecell{SBERT-Comp.\\(w/o validation)}} & \multicolumn{2}{c|}{\makecell{SBERT-Comp.\\(w/ validation)}}\\ \cline{2-9}
        & \makecell{$\rho$} & \makecell{$r_s$} & \makecell{$\rho$} & \makecell{$r_s$} & \makecell{$\rho$} & \makecell{$r_s$} & \makecell{$\rho$} & \makecell{$r_s$}\\ \hline

        appceleratorstudio & 0.3254 & 0.3037 & 0.2904 & 0.2786 & \textbf{0.3330} & \textbf{0.3222} & 0.2958 & 0.2866\\
        aptanastudio & 0.3419 & \textbf{0.2830} & 0.2497 & 0.1680 & \textbf{0.3452} & 0.2682 & 0.3027 & 0.2400\\
        bamboo & 0.1768 & 0.1753 & \textbf{0.2516} & \textbf{0.2114} & 0.1860 & 0.1761 & 0.0876 & 0.0953\\
        clover & \textbf{0.4403} & 0.4166 & 0.4059 & 0.3834 & 0.4190 & \textbf{0.4483} & 0.4190 & 0.4006\\
        datamanagement & \textbf{0.3775} & 0.3909 & 0.3389 & \textbf{0.3974} & 0.3271 & \textbf{0.3794} & 0.3021 & 0.3864\\
        duracloud & 0.3758 & \textbf{0.4221} & 0.3593 & 0.3792 & \textbf{0.3858} & 0.4006 & 0.3829 & 0.3939\\
        jirasoftware & \textbf{0.5324} & \textbf{0.4987} & 0.4522 & 0.4463 & 0.4414 & 0.4386 & 0.4915 & 0.4442\\
        mesos & 0.3960 & 0.3916 & 0.4156 & 0.4218 & \textbf{0.4359} & \textbf{0.4402} & 0.4053 & 0.4271\\
        moodle & \textbf{0.3499} & 0.3552 & 0.2794 & 0.3169 & 0.2929 & 0.3276 & 0.3301 & \textbf{0.3574}\\
        mule & 0.2342 & 0.2459 & 0.2209 & 0.2392 & \textbf{0.3188} & \textbf{0.3235} & 0.2306 & 0.2498\\
        mulestudio & 0.1096 & 0.0612 & 0.1616 & 0.1654 & 0.2265 & 0.2148 & \textbf{0.2605} & \textbf{0.2413}\\
        springxd & 0.3982 & 0.3911 & 0.3888 & 0.3867 & \textbf{0.4066} & 0.4041 & 0.3984 & \textbf{0.4155}\\
        talenddataquality & 0.2892 & 0.2985 & \textbf{0.3015} & 0.2949 & 0.2983 & \textbf{0.2973} & 0.2363 & 0.2393\\
        talendesb & 0.3453 & 0.3550 & 0.4007 & 0.4180 & \textbf{0.4189} & 0.4528 & 0.4165 & \textbf{0.4567}\\
        titanium & 0.1861 & 0.2264 & 0.1828 & \textbf{0.2519} & \textbf{0.2098} & 0.2439 & 0.1945 & 0.1881\\
        usergrid & 0.2016 & 0.1981 & 0.2895 & 0.2764 & 0.2945 & \textbf{0.3075} & \textbf{0.3020} & 0.3059\\ \hline
        Average & 0.3175 & 0.3133 & 0.3118 & 0.3147 & \textbf{0.3337} & \textbf{0.3403} & 0.3160 & 0.3205\\ \hline

    \end{tabular}
    \label{table:pairwise_baselines}
\end{table*}
\vspace*{-4ex}

Table \ref{table:pairwise_baselines} compares the SBERT-Regression model trained on direct story points with models trained on comparative judgments, including SBERT-Comparative ( with and without validation data) and LinearSVM-Comparative from Qian et al. \cite{qian2015learning}. All the results of the comparative models in Table~\ref{table:pairwise_baselines} were trained with $k=1$ pair for each training data point to allow a fair comparison with the regression models. From Table \ref{table:pairwise_baselines} we observe:
\bi
\item SBERT-Comparative outperformed LinearSVM-Comparative in most projects due to their flexibility in the encoder architecture. Only in the bamboo project did LinearSVM-Comparative achieve a better performance.
\item SBERT-Comparative (without validation) outperformed SBERT-Comparative (with validation) in most projects, except for moodle and mulestudio, due to its access to richer information in the union of training and validation sets.
\item SBERT-Comparative (with or without validation) achieved better performance than SBERT-Regression in many projects and similar performance in other projects. The regression model performed better only in the jirasoftware project.
\item In general, SBERT-Comparative (without validation) showed the best performance in either $\rho$ or $r_s$ in 12 out of 16 projects, featuring the highest average $\rho=0.33$ and $r_s=0.34$. \\
\ei

% \begin{RQ}
\rules{\noindent\textbf{Answer to RQ2: }Models learning from comparative judgments can perform similarly to, if not better than, models trained on direct ratings. Since the training size of comparative judgments is equal to the training size of direct ratings, less human annotation effort is required for comparative judgments (according to the law of comparative judgment).}
% \end{RQ}

\subsubsection{RQ3 Do more training data pairs lead to better performance?}
\label{sec:RQ3}
\begin{figure}[h!]
    \centering
    % \hspace{-1cm}
    \includegraphics[width=0.85\linewidth]{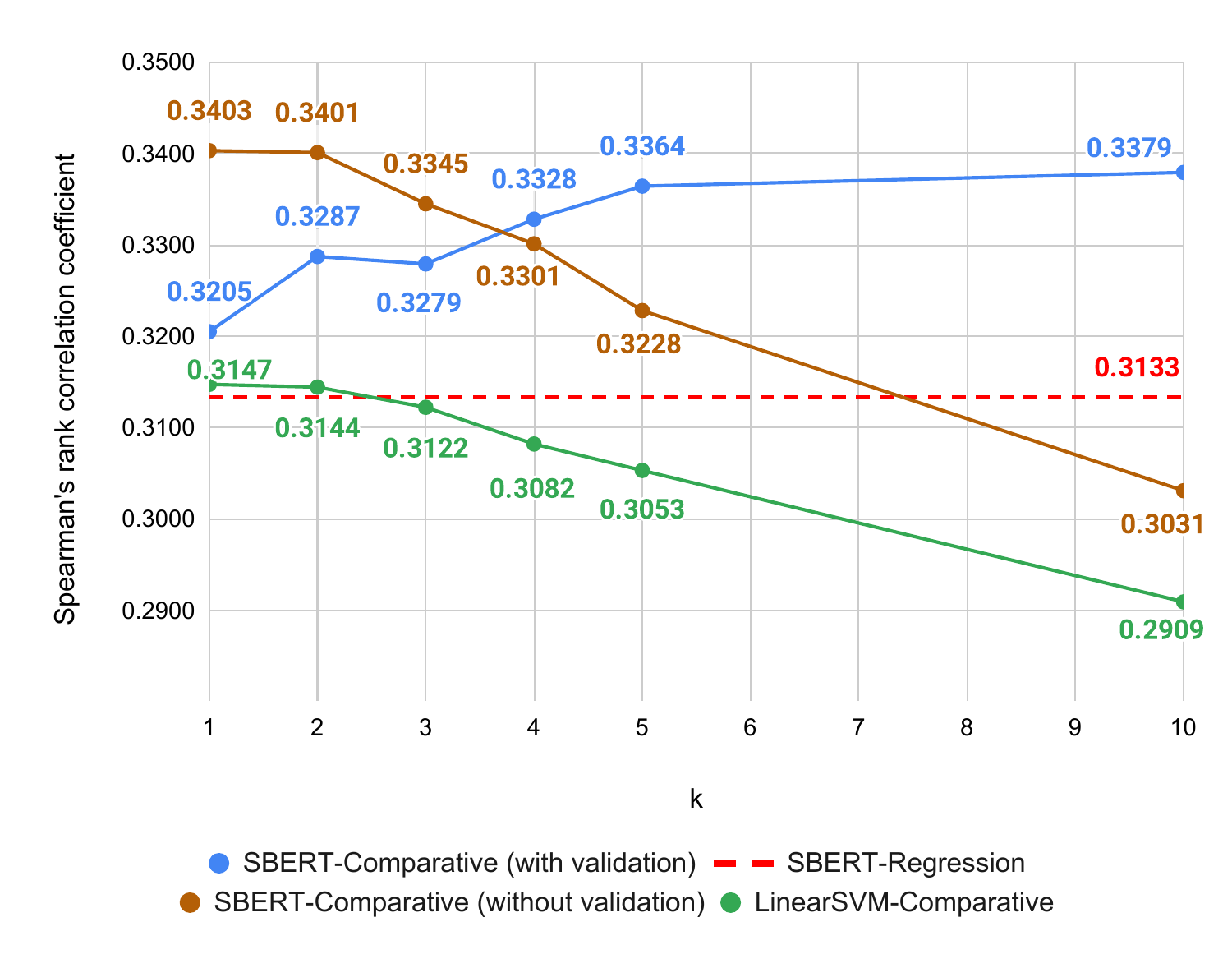}
    \caption{Results for RQ3.}
    \label{figure:pair_vs_spearmans}
\end{figure}
As shown in Table \ref{table:pairwise_baselines}, our approach achieved better ranking performance than the baseline LinearSVM-Comparative model when trained on $n$ comparative judgments. However, up to $n^2$ comparative judgments are available for a backlog of $n$ items. Thus, we evaluate in this research question whether increasing the number of comparative judgments yields better prediction performance. 

% \vspace{-1ex}
Figure \ref{figure:pair_vs_spearmans} shows the average $r_s$ of each comparative model with $k \in \{1,2,3,4,5,10\}$ data pairs generated for each training data point. This makes the total number of training data pairs to be $(k \times n)$.

\bi
\item
All comparative models perform similarly to or better than the regression model SBERT-Regression when $k$ is small. This validates the findings of RQ2.
\item
For comparative models without validation, the performance worsens with a larger $k$. This suggests overfitting of the larger training data.
\item
SBERT-Comparative (with validation) benefits from a larger $k$ because of its ability to prevent overfitting with a validation set. However, the results also suggest diminishing returns for it: increasing the number of comparative judgments only marginally increases the prediction performance for $k\ge 5$.\\
\ei

\rules{\noindent\textbf{Answer to RQ3: }Comparative learning models can benefit from more training data pairs when validation is utilized to prevent overfitting. However, the best performance was achieved by training SBERT-Comparative on a more diverse dataset $\mathcal{D}_{train} \cup \mathcal{D}_{val}$ with minimal annotation cost.}

\section{Cross-project experiments}\label{sect:cross}
This section focuses on the following research question:\\
\begin{itemize}
    \item \textbf{RQ4 How does the proposed model perform in cross-project predictions?} We conducted similar experiments as for \textbf{RQ1} and \textbf{RQ2}, but with a different configuration of the data for these cross-project experiments. These experiments involved training each model on all but one of the 16 projects' training sets to test the entire data of the remaining project.
\end{itemize}

\subsection{Experimental setup}
These experiments followed the same overall process as the machine learning experiments. The only difference is the dataset configuration. As discussed, for each dataset, we held the entire dataset with all three splits as a larger testing set. We used the training set of all the remaining projects as a larger training set and followed a similar process for the validation set. The same evaluation metrics used for RQ1-RQ4 were used to evaluate these experiments.

% \yiming{Tables are too wide.}
\subsection{Results}
Table-\ref{table:issue_baselines_cross} and Table-\ref{table:pairwise_baselines_cross} present the results of these cross-project experiments. We noticed a decrease in the values of all metrics for both the proposed and state-of-the-art models. This is expected, as different projects have different patterns, vocabularies, and syntaxes, which can lead to poorer learning unless the model is built for such diverse data. Nonetheless, our proposed regression model outperformed state-of-the-art models for the regression task. Our proposed comparative models also exhibited comparable performance for the comparative task.

However, the overall performance in cross-project settings is much worse than that in within-project settings for both our proposed and state-of-the-art models. This indicates that the use of machine learning models for story point estimation in cross-project settings may not be viable. Therefore, project-specific training is a better choice for automated story point estimation.\\

\rules{ \noindent\textbf{Answer to RQ4: }The results indicate the our proposed models trained and evaluated in a cross-project setting can show comparable performance to state-of-the-art models for both the regression and comparative tasks. However, cross-project prediction performance is still much worse than within-project prediction performance. Therefore, project-specific training data might still be necessary.}

\begin{table*}[!thp]
    \centering
    \renewcommand{\arraystretch}{1.25}
    % \begin{tiny}
    \caption{Results for RQ4 (Regression)}
    \begin{tabular}{|c|c|c|c||c|c|c||c|c|c|}
    \hline
        \multirow{2}{*}{Project} & \multicolumn{3}{c||}{\makecell{FastText-SVM \cite{atoum2024enhancing}}} & \multicolumn{3}{c||}{GPT2SP \cite{fu2022gpt2sp}} & \multicolumn{3}{c|}{\textbf{\makecell{SBERT-\\Regression}}}\\ \cline{2-10}
        
        & \makecell{$\rho$} & \makecell{$r_s$} & MAE & \makecell{$\rho$} & \makecell{$r_s$} & MAE & \makecell{$\rho$} & \makecell{$r_s$} & MAE\\ \hline

        appceleratorstudio & 0.13 & 0.13 & 2.48 & 0.16 & 0.15 & 1.86 & 0.19 & 0.18 & 2.50\\
        aptanastudio & 0.23 & 0.26 & 4.77 & 0.17 & 0.15 & 1.86 & 0.27 & 0.29 & 4.49\\
        bamboo & 0.14 & 0.13 & 2.07 & 0.20 & 0.19 & 1.83 & 0.15 & 0.20 & 1.91\\
        clover & 0.28 & 0.23 & 3.49 & 0.20 & 0.16 & 1.87 & 0.32 & 0.26 & 3.60\\
        datamanagement & 0.13 & 0.27 & 7.49 & 0.17 & 0.19 & 1.83 & 0.20 & 0.34 & 7.41\\
        duracloud & 0.14 & 0.16 & 2.67 & 0.20 & 0.16 & 1.84 & 0.24 & 0.22 & 2.71\\
        jirasoftware & 0.18 & 0.23 & 2.37 & 0.24 & 0.23 & 1.80 & 0.08 & 0.15 & 2.79\\
        mesos & 0.25 & 0.28 & 1.81 & 0.25 & 0.22 & 1.88 & 0.29 & 0.35 & 1.91\\
        moodle & 0.21 & 0.23 & 12.55 & 0.24 & 0.21 & 1.88 & 0.26 & 0.34 & 12.21\\
        mule & 0.23 & 0.20 & 2.87 & 0.24 & 0.23 & 1.86 & 0.28 & 0.26 & 2.69\\
        mulestudio & 0.18 & 0.22 & 3.40 & 0.19 & 0.18 & 1.86 & 0.24 & 0.24 & 3.38\\
        springxd & 0.26 & 0.29 & 2.15 & 0.18 & 0.16 & 2.32 & 0.27 & 0.29 & 2.11\\
        talenddataquality & 0.26 & 0.29 & 3.57 & 0.33 & 0.33 & 1.65 & 0.29 & 0.33 & 3.54\\
        talendesb & 0.38 & 0.41 & 2.24 & 0.20 & 0.19 & 2.09 & 0.37 & 0.40 & 1.88\\
        titanium & 0.19 & 0.19 & 3.34 & 0.06 & 0.04 & 1.48 & 0.25 & 0.21 & 3.32\\
        usergrid & 0.21 & 0.15 & 1.61 & 0.11 & 0.11 & 2.07 & 0.30 & 0.26 & 1.56\\ \hline
        Average & 0.21 & 0.23 & 3.68 & 0.20 & 0.18 & 1.87 & 0.25 & 0.27 & 3.63\\
        \hline
        
    \end{tabular}
% \end{tiny}
    \label{table:issue_baselines_cross}
\end{table*}

\begin{table*}[!thp]
    \centering
    \renewcommand{\arraystretch}{1.25}
    \caption{Results for RQ4 (Comparative)}
    \begin{tabular}{|c||c|c||c|c||c|c||c|c|}
    \hline
        \multirow{2}{*}{Project} & \multicolumn{2}{c||}{\makecell{SBERT-\\ Regression}} & \multicolumn{2}{c||}{\makecell{LinearSVM-\\Comp.}} & \multicolumn{2}{c||}{\makecell{SBERT-Comp.\\ (w/o validation)}} & \multicolumn{2}{c|}{\makecell{SBERT-Comp.\\ (w/ validation)}}\\ \cline{2-9}
        & \makecell{$\rho$} & \makecell{$r_s$} & \makecell{$\rho$} & \makecell{$r_s$} & \makecell{$\rho$} & \makecell{$r_s$} & \makecell{$\rho$} & \makecell{$r_s$}\\ \hline

        appceleratorstudio & 0.19 & 0.18 & 0.19 & 0.18 & 0.19 & 0.19 & 0.19 & 0.18\\
        aptanastudio & 0.27 & 0.29 & 0.27 & 0.28 & 0.28 & 0.29 & 0.27 & 0.29\\
        bamboo & 0.15 & 0.20 & 0.16 & 0.19 & 0.16 & 0.19 & 0.16 & 0.18\\
        clover & 0.32 & 0.26 & 0.31 & 0.25 & 0.31 & 0.24 & 0.29 & 0.24\\
        datamanagement & 0.20 & 0.34 & 0.20 & 0.32 & 0.18 & 0.32 & 0.20 & 0.33\\
        duracloud & 0.24 & 0.22 & 0.22 & 0.20 & 0.21 & 0.20 & 0.25 & 0.24\\
        jirasoftware & 0.08 & 0.15 & 0.06 & 0.13 & 0.05 & 0.12 & 0.08 & 0.17\\
        mesos & 0.29 & 0.35 & 0.30 & 0.36 & 0.30 & 0.36 & 0.30 & 0.36\\
        moodle & 0.26 & 0.34 & 0.28 & 0.38 & 0.27 & 0.37 & 0.29 & 0.38\\
        mule & 0.28 & 0.26 & 0.30 & 0.27 & 0.29 & 0.26 & 0.26 & 0.23\\
        mulestudio & 0.24 & 0.24 & 0.24 & 0.25 & 0.23 & 0.24 & 0.24 & 0.24\\
        springxd & 0.27 & 0.29 & 0.28 & 0.30 & 0.28 & 0.30 & 0.29 & 0.32\\
        talenddataquality & 0.29 & 0.33 & 0.25 & 0.27 & 0.24 & 0.25 & 0.24 & 0.23\\
        talendesb & 0.37 & 0.40 & 0.37 & 0.40 & 0.36 & 0.39 & 0.37 & 0.39\\
        titanium & 0.25 & 0.21 & 0.23 & 0.19 & 0.23 & 0.20 & 0.19 & 0.16\\
        usergrid & 0.30 & 0.26 & 0.27 & 0.23 & 0.26 & 0.21 & 0.26 & 0.20\\ \hline
        Average & 0.25 & 0.27 & 0.24 & 0.26 & 0.24 & 0.26 & 0.24 & 0.26\\
        
        \hline

    \end{tabular}
    \label{table:pairwise_baselines_cross}
\end{table*}

%%%%%%%%%%%%%%%%%%%%%%%%%%%%%%%%%%%%%%%%%%%%%%%%%%%%%%%%%%%%%%%%%%%%%%%%%%%%%%%%%%%
\newpage
%%%%%%%%%%%%%%%%%%%%%%%%%%%%%%%%%%%%%%%%%%%%%%%%%%%%%%%%%%%%%%%%%%%%%%%%%%%%%%%%%%%
\section{Human subject experiments}
\label{sec:human_exp}
Human subject experiments and their results were used to answer the final research questions. In this section, different accuracy, agreement, and ranking metrics were used alongside the human participant responses denoting their confidence in their responses. To this end, the following research questions were explored:\\
\begin{itemize}
    \item \textbf{RQ5 Is it faster and easier for human developers to provide the comparative judgments than direct story point estimates?}\\
    \item \textbf{RQ6 For story point estimation, are comparative judgments from human judges more reliable and of higher quality than direct story point estimates?} \\
\end{itemize}
Machine learning experiments work on the assumption that comparative approaches exert less cognitive effort from humans and require less time. This section describes the human subject experiments conducted to explore the validity of this assumption. The following subsections explain the design and results of these human subject experiments involving direct and comparative judgments on backlog items.

\subsection{Studies overview}
We conducted two experiments as a part of this human subject experiment section. These experiments are as follows - 
\begin{itemize}
    \item \textbf{HSE-1:} This experiment involved asking experienced software developers to provide their direct and comparative judgments on two different sets of user stories gathered from the JIRA project in our machine learning experiments.
    \item \textbf{HSE-2:} This experiment involved asking groups of students to provide their direct and comparative judgments both individually and as a group on a set of user stories generated by themselves as a part of their course project.
\end{itemize}

\subsection{HSE-1 study design}
In contrast to most story point estimation methods where it is a collaborative process between multiple members of a team, the human subject experiments here attempt to look at the effort and time required by human judges for story point estimation on an individual basis. The study is broadly separated into two sections - direct ratings and comparative ratings. Each section contains 10 multiple choice questions.

\begin{figure}[!h]
    \centering
    \begin{subfigure}{0.47\linewidth}
        \includegraphics[width=\linewidth]{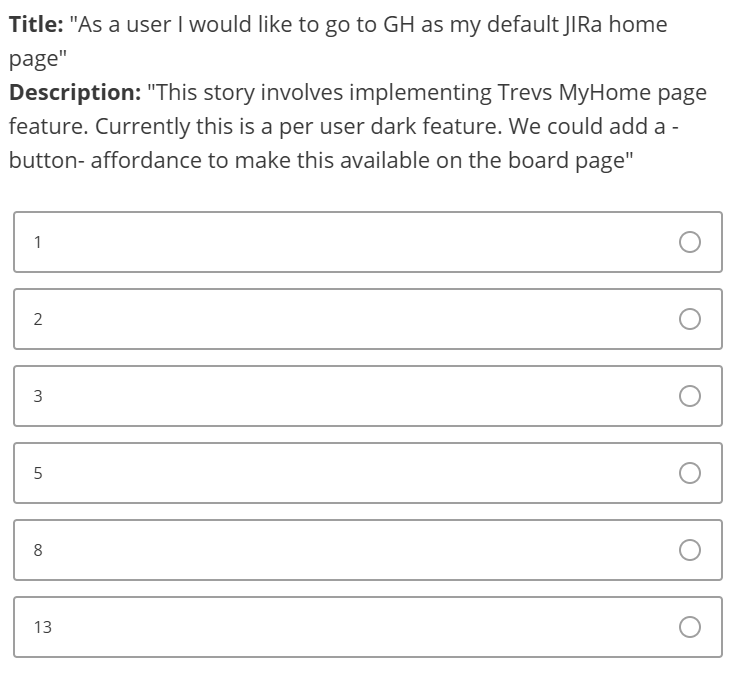}
        \caption{Sample question from the direct judgment section}
        \label{fig:sample_direct}
    \end{subfigure}
    ~ 
    \begin{subfigure}{0.47\linewidth}
        \includegraphics[width=\linewidth]{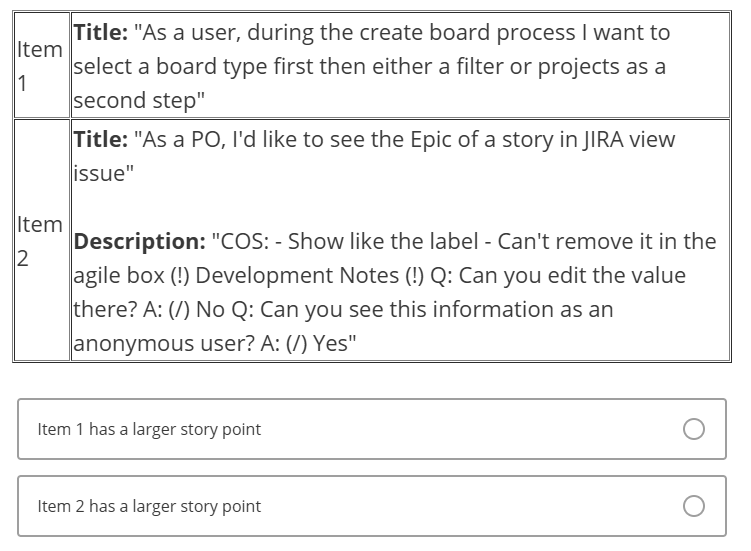}
        \caption{Sample question from the comparative judgment section}
        \label{fig:sample_comparative}
    \end{subfigure}
    \caption{Sample questions}
    \label{fig:sample_questions}
    \vspace*{-5ex}
\end{figure}

For the direct ratings section, each question has the text from some backlog item, and asks the participant to estimate the story point for this item. The participant can choose from multiple values listed under the question. These options are all the unique story point values from the original story points for these backlog items. Figure-\ref{fig:sample_direct} shows an example question from this section.

For the comparative ratings section, each question showcases the text from a pair of backlog items, and asks the participant to guess which item has the higher story point. The participants are presented with two options - ``Item 1 has a higher story point" and ``Item 2 has a higher story point". Figure-\ref{fig:sample_comparative} shows an example question from this section.

\begin{figure}[h!]
    \centering
    \begin{subfigure}{0.47\linewidth}
        \centering
        \includegraphics[width=\textwidth]{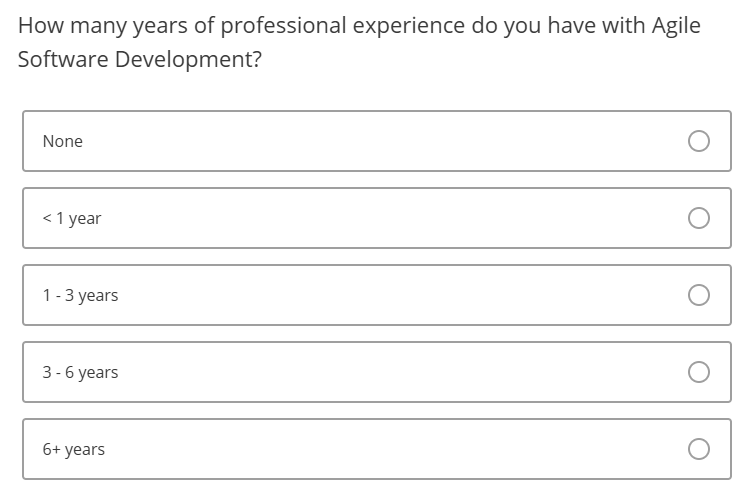}
        \caption{Sample experience question}
    \end{subfigure}%
    ~ 
    \begin{subfigure}{0.47\linewidth}
        \centering
        \includegraphics[width=\textwidth]{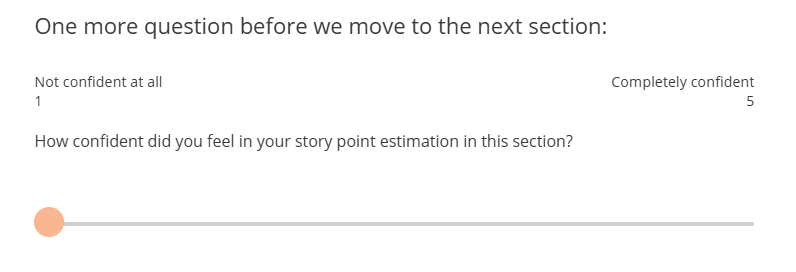}
        \caption{Sample confidence question}
    \end{subfigure}
    \caption{Sample optional questions}
    \label{fig:optional_questions}
\end{figure}

There are 3 additional optional questions before these two sections and one additional question after each of the two sections. The initial 3 optional questions ask the participant about their professional experience with story point estimation. The optional questions after the direct and comparative sections ask the participants how confident they felt in making these judgments. Figure-\ref{fig:optional_questions} shows how these optional questions look like to a participant. The 3 optional experience questions are as follows -

\begin{enumerate}
    \item How many years of professional experience do you have with Agile Software Development?
    \item How many years of professional experience do you have with Story point estimation?
    \item Have you ever taken part in Planning poker before?
\end{enumerate}

% \vspace*{-7ex}
\subsection{HSE-1 Data}
The story point estimation dataset introduced by Choetkiertikul et al. \cite{choetkiertikul2018deep} was used to pick each of the 10 questions for both the direct and comparative judgment sections. Each project was explored and jirasoftware was chosen to pick the items from, due to its prevalent use in software development lifecycles. After a pilot study involving a different set of items, the criteria decided on to pick the final set of items were as follows -\\

\begin{itemize}
    \item The text must start with ``As $<role>$, ..." since the pilot study participants achieved higher accuracy for items starting with this phase. With a clearer definition of whose perspective the item text is from or what the expectations are, it proved easier for participants to make accurate predictions.\\
    
    \item The item text must not be too long. While there were no strict limit placed on the number of words in each item text, the longest text contained about 80 words. Pilot study participants tended to provide more inaccurate predictions for longer items. It could have proved difficult for the participants to retain all the information in longer text, particularly when they were in pairwise fashion.\\
    
    \item No overlap in items between the direct and comparative sections. Having overlapping items could create the scenario where participants make predictions based on their previous predictions, creating a snowballing effect where past accurate or inaccurate predictions lead to later accurate or inaccurate predictions. Removing overlaps helped participants make predictions based solely on the item text presented in front of them.
\end{itemize}

\subsection{HSE-1 Participants}
Professionals with experience in software development lifecycles and story point estimation were reached out to participate in the study after obtaining informed consent. No personally identifiable information about the participants were collected. The participants' details can be seen in Table-\ref{table:participants}

\begin{table*}[!ht]
    \centering
    \renewcommand{\arraystretch}{1.25}
    \caption{Participants}
    \begin{tabular}{|c|c|c|}
    \hline
        Participant & \makecell{Professional\\ Experience} & \makecell{Story Point\\ Estimation Experience}\\ \hline
        P1 & 6+ years & 1 - 3 years\\ \hline
        P2 & 1 - 3 years & 1 - 3 years\\ \hline
        P3 & 3 - 6 years & 3 - 6 years\\ \hline
        P4 & $<$ 1 year & $<$ 1 year\\ \hline
        P5 & 1 - 3 years & $<$ 1 year\\
        \hline
    \end{tabular}
    \label{table:participants}
\end{table*}

\subsection{HSE-2 study design}
\label{sec:HSE2_study_design}
To more closely simulate story point estimation in a collective setting, we conducted further human subject experiments with groups of students going through story point estimation individually and as a group on a list of user story items generated by the groups themselves. These user stories are based on the group project they would be completing in the following months. Compared to the individual human subject experiments section, the participants here have less professional experience with story point estimation, but are going through this process on story items they are more familiar with, and are conducting it as a group as well as individually.
These group human subject experiments were comprised of the following steps -\\

\begin{itemize}
    \item Each team was asked to sit together for the remainder of this study, and have a copy of their user stories available at hand\\
    \item Each team assigned a numerical identifier among themselves\\
    \item Each team was split into two groups - Group A and Group B\\
    \item Participants in Group A were asked to individually assign story points to each user story ($I^p_{i}$ denotes the story point assigned to Item $i$ by Participant $p$) and log the total time\\
    \item Participants in Group B were asked to individually assign comparative labels to pre-selected pairs of user stories ($I^p_{ij}$ denotes the comparative judgment between Item $i$ and Item $j$ made by Participant $p$) and log the total time\\
    \item Participants in a team were asked to gather together to discuss and assign comparative labels to the same pairs of user stories ($G_{ij}$ denotes the consensus comparative judgment between Item $i$ and Item $j$) and log the total time\\
    \item Participants in a team were asked to gather together to discuss and assign story points to each user story ($G_{i}$ denotes the consensus story point assigned to Item $i$) and log the total time\\
    \item Participants were asked to answer four questions -\\
        \begin{itemize}
            \item \textbf{Question 1:} On a scale of 1 to 5 (with 1 being least confident and 5 being most confident), how confident do you feel with your group direct ratings?
            \item \textbf{Question 2:} On a scale of 1 to 5 (with 1 being least confident and 5 being most confident), how confident do you feel with your group comparative labels?
            \item \textbf{Question 3:} Do you feel that discussing the comparative labels together first had any positive impact on your direct rating afterwards? (Y/N)
            \item \textbf{Question 4:} Did you feel more confident in your group labels or in your individual labels or equally confident about both? (Group/Individual/Both)
        \end{itemize}
\end{itemize}

\subsection{HSE-2 Data}
Participants in each team used their own set of user stories based on their planned group project for the course. Teams were asked to make sure each team had at least 10 user stories. Teams 1-6 had 10, 11, 10, 13, 10 and 11 user stories respectively.
Which pre-selected pairs of user stories each team would work on was generated randomly based on the assumption of 10 user stories per team. These pairs were - $(10, 2), (5, 4), (3, 5), (6, 3), (8, 5),
(3, 4), (7, 4), (2, 9), (11, 5), (4, 2)$.

\subsection{HSE-2 Participants}
No personally identifiable information was recorded for the participants. Table-\ref{table:participants_RQ8} shows some information about the teams, their groups and data.

\begin{table*}[h!]
    \centering
    \renewcommand{\arraystretch}{1.25}
    \caption{Participants for RQ8}
    \begin{tabular}{|c|c|c|c|c|}
    \hline
    Team & Members & Group A & Group B & \makecell{Number of\\ user stories}\\
    \hline
    1 & 5 & 2, 4 & 1, 3, 5 & 10\\
    2 & 4 & 3, 4 & 1, 2 & 11\\
    3 & 5 & 1, 2 & 3, 4, 5 & 10\\
    4 & 4 & 1, 2 & 3, 4 & 13\\
    5 & 4 & 1, 2 & 3, 4 & 10\\
    6 & 5 & 3, 4 & 1, 2, 5 & 11\\
    \hline
    \end{tabular}
    \label{table:participants_RQ8}
\end{table*}
\vspace*{-6ex}
\subsection{Evaluation metrics}\label{sect:eval}

\subsubsection{RQ5}

The time required to make a prediction for each question (HSE-1) or the average time required to make a judgment in each section (HSE-2) was recorded. This information helped provide an insight into how long it took on average each human judge to make direct or comparative predictions. Finally, each participant was also asked to rate their confidence in their own predictions for both the direct and comparative sections on a scale of 1 to 5.

\subsubsection{RQ6}

The direct estimates from both HSE-1 and HSE-2 were first evaluated for their (1) inter-rater reliability, the $\rho$ and $r_s$ between each two raters (e.g. $\rho (I_i^p, I_i^q),\,\forall p\neq q$ in HSE-2); and (2) reliability against ground truth, the $\rho$ and $r_s$ between each rater and a set of ground truth story points (e.g. $\rho (I_i^p, G_i),\,\forall p$ in HSE-2). For HSE-1, the ground truth came from the original dataset~\cite{choetkiertikul2018deep} while for HSE-2, the ground truth was the final consensus estimates $G_i$.

On the other hand, the comparative judgments from both HSE-1 and HSE-2 were evaluated for their (1) inter-rater reliability, the $p_o$ and $\kappa$ between each two raters (e.g. $p_o (I_{ij}^p, I_{ij}^q),\,\forall p\neq q$ in HSE-2); and (2) reliability against ground truth, the $p_o$ and $\kappa$ between each rater and a set of ground truth story points (e.g. $p_o (I_{ij}^p, G_{ij}),\,\forall p$ in HSE-2). For HSE-1, the ground truth came from the original dataset~\cite{choetkiertikul2018deep} $y_{ij}=\text{sgn}(y_i-y_j)$ while for HSE-2, the ground truth was the final consensus estimates $G_{ij}$. Here, $p_o$ is the observed agreement
\begin{equation}
    p_o = \cfrac{m}{n}
\end{equation}
where $n$ is the total number of comparative judgments and $m$ is the number of agreements (e.g. when $I_{ij}^p = I_{ij}^q$). The Cohen's $\kappa$  coefficient is calculated as 
\begin{equation}
    \kappa = 1 - \cfrac{1 - p_o}{1 - p_e}
\end{equation}
where $p_e$ denotes the hypothetical probability of chance agreement
\begin{equation}
    p_e = \cfrac{1}{n^2}{\sum_k}n_{k1}n_{k2}
\end{equation}
where, $k=2$ is the number of labels a participant could assign to a single item pair (1 or -1), and $n_{k1}$ and $n_{k2}$ are the number of labels assigned to category $k$ by the first and second participant respectively.

In order to directly compare the reliability of direct estimates and comparative judgments, we also generated potential comparative judgments with the direct estimates: $\hat{I}_{ij}^p = \text{sgn}(I_i^p-I_j^p)$. Using these generated comparative judgments, we can also evaluate the inter-rater reliability and reliability against ground truth of the direct estimates using $p_o$ and $\kappa$. This allows a direct comparison of the average $p_o$ and $\kappa$ achieved between direct estimates and comparative judgments. Note that, this is still not an apple-to-apple comparison since (1) for HSE-1, the estimates from the same raters were compared but the comparative judgments were made on different data items; and (2) for HSE-2, the comparative judgments on the same data item pairs were compared but they were provided by different human raters.

\subsection{Results}
\subsubsection{RQ5 Is it faster and easier for human developers to provide the comparative judgments than direct story point estimates?}

\begin{table*}[tbh!]
    \centering
    \renewcommand{\arraystretch}{1.25}
    \caption{Comparing participants' efficiency between direct and comparative judgments using their reported confidence levels and average time taken (in seconds) for each judgment (RQ5, HSE-1)}
    \begin{tabular}{|c|c|c|c|c|}
    \hline
    \multirow{2}{*}{Participant} & \multicolumn{2}{c|}{Direct} & \multicolumn{2}{c|}{Comparative}\\ \cline{2-5}
    & \makecell{Conf\\ (1-5)} & \makecell{Time\\ (s)} & \makecell{Conf\\ (1-5)} & \makecell{Time\\ (s)}\\ \hline
    P1 & 3 & 90.88 & 3 & 73.42\\
    P2 & 3 & 24.92 & 3 & 8.67\\
    P3 & 3 & 22.41 & 3 & 16.38\\
    P4 & 2 & 19.88 & 4 & 21.48\\
    P5 & 3 & 31.17 & 3 & 34.04\\ \hline
    Average & 2.8 & 37.85 & 3.2 & 30.80\\
    \hline
    \end{tabular}
    \label{table:human_quick}
\end{table*}

\begin{table*}[tbh!]
    \centering
    \caption{Comparing participants' efficiency between direct and comparative judgments using their reported confidence levels and average time taken (in seconds) for each judgment (RQ5, HSE-2), where ``I" denotes the judgments made individually by each participant and ``G" denotes the consensus judgments of the group.}
    \begin{tabular}{|c|c||c|c||c|c|}
    \hline
        \multirow{2}{*}{Group} & \multirow{2}{*}{Type} & \multicolumn{2}{c|}{Direct} & \multicolumn{2}{c|}{Comparative}\\ \cline{3-6}
        & & \makecell{Time\\ (s)} & \makecell{Conf\\ (1-5)} & \makecell{Time\\ (s)} & \makecell{Conf\\ (1-5)}\\ \hline
        \multirow{2}{*}{1} & I & 20.66 & 3.50 & 15.86 & 3.33\\
        & G & 41.51 & 4.20 & 75.05 & 3.20\\ \hline
        \multirow{2}{*}{2} & I & 10.97 & 3.00 & 17.07 & 4.00\\
        & G & 9.90 & 3.33 & 17.04 & 4.00\\ \hline
        \multirow{2}{*}{3} & I & 26.50 & 3.00 & 15.28 & 3.66\\
        & G & 71.00 & 3.40 & 45.00 & 3.80\\ \hline
        \multirow{2}{*}{4} & I & 9.24 & 4.50 & 8.91 & 4.00\\
        & G & 25.80 & 4.25 & 15.20 & 4.00\\ \hline
        \multirow{2}{*}{5} & I & 13.50 & 3.50 & 15.50 & 5.00\\
        & G & 42.18 & 3.75 & 22.00 & 4.50\\ \hline
        \multirow{2}{*}{6} & I & 10.22 & 4.00 & 12.42 & 4.33\\
        & G & 27.91 & 4.00 & 20.50 & 4.20\\ \hline
        \multirow{2}{*}{Average} & I & 15.18 & 3.58 & 14.17 & 4.05\\
        & G & 36.38 & 3.82 & 32.47 & 3.95\\
    \hline
    \end{tabular}
    \label{table:RQ5_HSE2}
\end{table*}

Table-\ref{table:human_quick} compares the confidences expressed by each participant in HSE-1 and the average time taken to make a judgment on an item (for the direct rating section) or a pair of items (for the comparative rating section). 
On average, the participants showed higher confidence for the comparative section while taking less time to make these decisions. These reported times include the time taken to read the text of item(s) as well, further emphasizing the ease of making comparative judgments, since that section requires reading the text of a pair of items, compared to reading the text from only one item in the direct rating section.

Similarly for HSE-2, participants showed more confidence and took less time for their comparative judgment sections, as seen in Table-\ref{table:RQ5_HSE2}. Here, ``I" denotes the average time taken per participant during their individual sections for each item's estimation, while ``G" represents the time it took for that group to reach a consensus on the labels for each pair on average. 
On average, teams show slightly higher confidence for comparative story point estimation ($3.95$) over direct story point estimation ($3.82$). Teams also take less time on average ($32.47s$) to estimate the label for each pair of user stories compared to how long they take ($36.38s$) to estimate the story point for each user story directly. This time includes the time to read the user story or the pair of user stories, making the time difference more significant.\\

% Overall, participants showed more confidence and taking less time for comparative judgments, compared to direct judgments. Therefore, comparative judgments were observed to be more efficient than direct judgments for human judges.\\

\rules{ \noindent\textbf{Answer to RQ5: }Participants, on average, took less time yet showed more confidence in their comparative judgments than their direct judgments. In other words, participants showed more efficiency comparative judgments than direct judgments. }

\subsubsection{RQ6 For story point estimation, are comparative judgments more reliable than direct story point estimates?}

\noindent \textbf{HSE-1: }First, we analyze the inter-rater reliability of HSE-1. Table-\ref{table:raw_direct} and Table-\ref{table:raw_comparative} list the responses from the participants for the direct and comparative sections of HSE-1 respectively, alongside the ground truth scores or labels for each question.

\begin{table*}[h!]
    \centering
    \renewcommand{\arraystretch}{1.25}
    \caption{Responses and ground truth for the direct estimates (HSE-1)}
    \begin{tabular}{|c|c|c|c|c|c|c|c|c|c|c|c|}
    \hline
    Participant & Q1 & Q2 & Q3 & Q4 & Q5 & Q6 & Q7 & Q8 & Q9 & Q10 & Confidence\\ \hline
    P1 & 5 & 13 & 3 & 5 & 13 & 3 & 8 & 13 & 13 & 8 & 3\\
    P2 & 3 & 8 & 3 & 2 & 2 & 1 & 2 & 3 & 3 & 2 & 3\\
    P3 & 1 & 3 & 1 & 2 & 2 & 2 & 2 & 2 & 2 & 1 & 3\\
    P4 & 1 & 3 & 2 & 2 & 3 & 1 & 3 & 1 & 3 & 1 & 2\\
    P5 & 5 & 5 & 2 & 2 & 3 & 2 & 3 & 2 & 3 & 2 & 3\\ \hline
    Ground truth & 1 & 13 & 2 & 8 & 5 & 2 & 8 & 5 & 3 & 5 & -\\
    \hline
    \end{tabular}
    \label{table:raw_direct}
\end{table*}
% \vspace*{-3ex}
\begin{table*}[h!]
    \centering
    % \vspace*{-2ex}
    \renewcommand{\arraystretch}{1.25}
    \caption{Responses and ground truth for the comparative judgments (HSE-1)}
    \begin{tabular}{|c|c|c|c|c|c|c|c|c|c|c|c|}
    \hline
    \makecell{Participant} & Q1 & Q2 & Q3 & Q4 & Q5 & Q6 & Q7 & Q8 & Q9 & Q10 & \makecell{Confidence}\\ \hline
    P1 & 1 & -1 & 1 & -1 & 1 & 1 & 1 & 1 & 1 & -1 & 3\\
    P2 & -1 & -1 & 1 & -1 & 1 & 1 & -1 & 1 & -1 & -1 & 3\\
    P3 & 1 & -1 & 1 & -1 & 1 & 1 & 1 & 1 & -1 & -1 & 3\\
    P4 & 1 & -1 & -1 & 1 & 1 & 1 & -1 & 1 & -1 & -1 & 4\\
    P5 & 1 & -1 & -1 & -1 & 1 & 1 & -1 & 1 & -1 & 1 & 3\\ \hline
    % \makecell{Direct\\ scores} & 5 \& 1 & 2 \& 8 & 8 \& 13 & 5 \& 8 & 5 \& 2 & 3 \& 2 & 3 \& 13 & 13 \& 1 & 8 \& 3 & 1 \& 8 & -\\ \hline
    \makecell{Ground truth} & 1 & -1 & -1 & -1 & 1 & 1 & -1 & 1 & 1 & -1 & -\\
    \hline
    \end{tabular}
    \label{table:raw_comparative}
\end{table*}

Using the responses from Table~\ref{table:raw_direct}, Table-\ref{table:direct_eval} evaluates the participants' responses against each other and against the ground truth using the $\rho$, $r_s$ and $MAE$ metrics. These results are not directly comparable to the SBERT-Regression model's results since the human subject experiments only use a very small subset of the data compared to SBERT-Regression. However, the values being similar to those experiments is a promising sign for the model's potential use instead of relying on human labeling.

\begin{table*}[h!]
    \centering
    \caption{Direct judgment section response evaluation for each participant (HSE-1) using Pearson's Correlation Coefficient ($\rho$), Spearman's Rank Correlation Coefficient ($r_s$) and Mean Average Error ($MAE$) scores.}
    \begin{tabular}{|c|c|c|c|c|c|c|c|}
    \hline
        Participant & Metric & P1 & P2 & P3 & P4 & P5 & \makecell{Ground\\ truth}\\ \hline
        {\multirow{3}{*}{P1}} & $\rho$ & - & 0.451 & 0.564 & 0.493 & 0.268 & 0.441\\
         & $r_s$ & - & 0.397 & 0.542 & 0.487 & 0.418 & 0.502\\
         & $MAE$ & - & 5.5 & 6.6 & 6.4 & 5.5 & 3.8\\ \hline
        {\multirow{3}{*}{P2}} & $\rho$ & 0.451 & - & 0.533 & 0.370 & 0.675 & 0.641\\
         & $r_s$ & 0.397 & - & 0.100 & 0.208 & 0.481 & 0.036\\
         & $MAE$ & 5.5 & - & 1.3 & 1.3 & 1 & 2.9\\ \hline
        {\multirow{3}{*}{P3}} & $\rho$ & 0.564 & 0.533 & - & 0.559 & 0.264 & 0.742\\
         & $r_s$ & 0.542 & 0.100 & - & 0.557 & 0.279 & 0.668\\
         & $MAE$ & 6.6 & 1.3 & - & 0.6 & 1.1 & 3.4\\ \hline
        {\multirow{3}{*}{P4}} & $\rho$ & 0.493 & 0.370 & 0.559 & - & 0.295 & 0.517\\
         & $r_s$ & 0.487 & 0.208 & 0.557 & - & 0.507 & 0.515\\
         & $MAE$ & 6.4 & 1.3 & 0.6 & - & 0.9 & 3.2\\ \hline
        {\multirow{3}{*}{P5}} & $\rho$ & 0.268 & 0.675 & 0.264 & 0.295 & - & 0.285\\
         & $r_s$ & 0.418 & 0.481 & 0.279 & 0.507 & - & 0.107\\
         & $MAE$ & 5.5 & 1 & 1.1 & 0.9 & - & 3.1\\
    \hline
    \end{tabular}
    \label{table:direct_eval}
\end{table*}

\begin{table*}[!h]
    % \tiny
    \centering
    \renewcommand{\arraystretch}{1.15}
    \caption{Comparing the reliability between direct and comparative judgments (RQ6, HSE-1) using observed agreement ($p_o$) and Cohen's $\kappa$.}
    % \hspace*{-1cm}
    \begin{tabular}{|c|c|c|cccccc|}
\hline
\multirow{2}{*}{Metric}   & \multirow{2}{*}{\makecell{Parti-\\cipant}} & \multirow{2}{*}{Treatment} & \multicolumn{6}{c|}{Participant} \\ \cline{4-9} 
       &                          &                            & \multicolumn{1}{c|}{P1}     & \multicolumn{1}{c|}{P2}    & \multicolumn{1}{c|}{P3}     & \multicolumn{1}{c|}{P4}    & \multicolumn{1}{c|}{P5}     & \makecell{Ground\\ Truth} \\ \hline
\multirow{12}{*}{$p_o$}    & \multirow{2}{*}{P1}          & Direct                     & \multicolumn{1}{c|}{-}      & \multicolumn{1}{c|}{0.73}  & \multicolumn{1}{c|}{0.86}   & \multicolumn{1}{c|}{0.77}  & \multicolumn{1}{c|}{0.75}   & 0.7                     \\
                          &                              & Comparative                & \multicolumn{1}{c|}{-}      & \multicolumn{1}{c|}{0.7}   & \multicolumn{1}{c|}{0.9}    & \multicolumn{1}{c|}{0.6}   & \multicolumn{1}{c|}{0.6}    & 0.8                     \\ \cline{2-9} 
         & \multirow{2}{*}{P2}          & Direct                     & \multicolumn{1}{c|}{0.73}   & \multicolumn{1}{c|}{-}     & \multicolumn{1}{c|}{0.55}   & \multicolumn{1}{c|}{0.64}  & \multicolumn{1}{c|}{0.82}   & 0.5                     \\
                          &                              & Comparative                & \multicolumn{1}{c|}{0.7}    & \multicolumn{1}{c|}{-}     & \multicolumn{1}{c|}{0.55}   & \multicolumn{1}{c|}{0.64}  & \multicolumn{1}{c|}{0.82}   & 0.7                     \\ \cline{2-9} 
         & \multirow{2}{*}{P3}          & Direct                     & \multicolumn{1}{c|}{0.86}   & \multicolumn{1}{c|}{0.55}  & \multicolumn{1}{c|}{-}      & \multicolumn{1}{c|}{0.89}  & \multicolumn{1}{c|}{0.7}    & 0.92                    \\
                          &                              & Comparative                & \multicolumn{1}{c|}{0.9}    & \multicolumn{1}{c|}{0.55}  & \multicolumn{1}{c|}{-}      & \multicolumn{1}{c|}{0.89}  & \multicolumn{1}{c|}{0.7}    & 0.7                     \\ \cline{2-9} 
         & \multirow{2}{*}{P4}          & Direct                     & \multicolumn{1}{c|}{0.77}   & \multicolumn{1}{c|}{0.64}  & \multicolumn{1}{c|}{0.89}   & \multicolumn{1}{c|}{-}     & \multicolumn{1}{c|}{0.8}    & 0.79                    \\
                          &                              & Comparative                & \multicolumn{1}{c|}{0.6}    & \multicolumn{1}{c|}{0.64}  & \multicolumn{1}{c|}{0.89}   & \multicolumn{1}{c|}{-}     & \multicolumn{1}{c|}{0.8}    & 0.8                     \\ \cline{2-9} 
     & \multirow{2}{*}{P5}          & Direct                     & \multicolumn{1}{c|}{0.75}   & \multicolumn{1}{c|}{0.82}  & \multicolumn{1}{c|}{0.7}    & \multicolumn{1}{c|}{0.8}   & \multicolumn{1}{c|}{-}      & 0.57                    \\
                          &                              & Comparative                & \multicolumn{1}{c|}{0.6}    & \multicolumn{1}{c|}{0.82}  & \multicolumn{1}{c|}{0.7}    & \multicolumn{1}{c|}{0.8}   & \multicolumn{1}{c|}{-}      & 0.8                     \\ \cline{2-9} 
     & \multirow{2}{*}{\textbf{\makecell{Ave-\\ rage}}}     & Direct                     & \multicolumn{1}{c|}{0.7775} & \multicolumn{1}{c|}{0.685} & \multicolumn{1}{c|}{0.75}   & \multicolumn{1}{c|}{0.775} & \multicolumn{1}{c|}{0.7675} & 0.696                   \\ 
                       &                              & Comparative                & \multicolumn{1}{c|}{0.7}    & \multicolumn{1}{c|}{0.725} & \multicolumn{1}{c|}{0.775}  & \multicolumn{1}{c|}{0.7}   & \multicolumn{1}{c|}{0.7}    & 0.76                    \\ \hline 
\multirow{12}{*}{$\kappa$} & \multirow{2}{*}{P1}          & Direct                     & \multicolumn{1}{c|}{-}      & \multicolumn{1}{c|}{0.5}   & \multicolumn{1}{c|}{0.71}   & \multicolumn{1}{c|}{0.54}  & \multicolumn{1}{c|}{0.53}   & 0.35                    \\
                          &                              & Comparative                & \multicolumn{1}{c|}{-}      & \multicolumn{1}{c|}{0.44}  & \multicolumn{1}{c|}{0.78}   & \multicolumn{1}{c|}{0.2}   & \multicolumn{1}{c|}{0.2}    & 0.6                     \\ \cline{2-9} 
         & \multirow{2}{*}{P2}          & Direct                     & \multicolumn{1}{c|}{0.5}    & \multicolumn{1}{c|}{-}     & \multicolumn{1}{c|}{0.55}   & \multicolumn{1}{c|}{0.64}  & \multicolumn{1}{c|}{0.82}   & 0                       \\
                          &                              & Comparative                & \multicolumn{1}{c|}{0.44}   & \multicolumn{1}{c|}{-}     & \multicolumn{1}{c|}{0.62}   & \multicolumn{1}{c|}{0.4}   & \multicolumn{1}{c|}{0.4}    & 0.4                     \\ \cline{2-9} 
         & \multirow{2}{*}{P3}          & Direct                     & \multicolumn{1}{c|}{0.71}   & \multicolumn{1}{c|}{0.55}  & \multicolumn{1}{c|}{-}      & \multicolumn{1}{c|}{0.89}  & \multicolumn{1}{c|}{0.7}    & 0.83                    \\
                          &                              & Comparative                & \multicolumn{1}{c|}{0.78}   & \multicolumn{1}{c|}{0.62}  & \multicolumn{1}{c|}{-}      & \multicolumn{1}{c|}{0.4}   & \multicolumn{1}{c|}{0.4}    & 0.4                     \\ \cline{2-9} 
         & \multirow{2}{*}{P4}          & Direct                     & \multicolumn{1}{c|}{0.54}   & \multicolumn{1}{c|}{0.64}  & \multicolumn{1}{c|}{0.89}   & \multicolumn{1}{c|}{-}     & \multicolumn{1}{c|}{0.8}    & 0.57                    \\
                          &                              & Comparative                & \multicolumn{1}{c|}{0.2}    & \multicolumn{1}{c|}{0.4}   & \multicolumn{1}{c|}{0.4}    & \multicolumn{1}{c|}{-}     & \multicolumn{1}{c|}{0.6}    & 0.6                     \\ \cline{2-9} 
         & \multirow{2}{*}{P5}          & Direct                     & \multicolumn{1}{c|}{0.53}   & \multicolumn{1}{c|}{0.82}  & \multicolumn{1}{c|}{0.7}    & \multicolumn{1}{c|}{0.8}   & \multicolumn{1}{c|}{-}      & 0.14                    \\
                          &                              & Comparative                & \multicolumn{1}{c|}{0.2}    & \multicolumn{1}{c|}{0.4}   & \multicolumn{1}{c|}{0.4}    & \multicolumn{1}{c|}{0.6}   & \multicolumn{1}{c|}{-}      & 0.6                     \\ \cline{2-9} 
       & \multirow{2}{*}{\textbf{\makecell{Ave-\\ rage}}}     & Direct                     & \multicolumn{1}{c|}{0.57}   & \multicolumn{1}{c|}{0.305} & \multicolumn{1}{c|}{0.4675} & \multicolumn{1}{c|}{0.54}  & \multicolumn{1}{c|}{0.4925} & 0.378                   \\
                          &                              & Comparative                & \multicolumn{1}{c|}{0.405}  & \multicolumn{1}{c|}{0.465} & \multicolumn{1}{c|}{0.55}   & \multicolumn{1}{c|}{0.4}   & \multicolumn{1}{c|}{0.4}    & 0.52                    \\ \hline
\end{tabular}
    \label{table:human_easy_2}
\end{table*}

Results in Table-\ref{table:raw_comparative} are used to calculate the observed agreement $p_o$ and the Cohen's $\kappa$ between each rater and against the ground truth for the comparative judgment annotations. As discussed in Section~\ref{sect:eval}, the items, ground truth values and provided ratings from Table-\ref{table:raw_direct} were also used to generate the observed agreement $p_o$ and the Cohen's $\kappa$ between each rater and against the ground truth for the direct estimates. 
Table-\ref{table:human_easy_2} showcases these results, where ${p_o}$ and $\kappa$ denote their relative observed agreement and $\kappa$ scores against each other and the ground truth, while the participants' self-reported confidence in their provided judgments for the direct and comparative sections can be seen in Table-\ref{table:raw_direct} and Table-\ref{table:raw_comparative} respectively.
These results show that participants generally provide more consistent ratings for the comparative section than the direct section, while also reporting feeling more confident about the comparative ratings than the direct ratings. The $\kappa$ scores being variable for the direct section but more consistent for the comparative section indicate that the participants' judgments were more consistent and reliable for the comparative section than the direct section. Specifically, the higher average $p_o$ and $\kappa$ of comparative judgments between each rater and the ground truth indicates a higher reliability of the comparative judgments than direct estimates.\\

% HSE-2
\noindent \textbf{HSE-2: }Similar to Table-\ref{table:direct_eval}, Table-\ref{table:direct_eval_HSE2} evaluates the participants' responses against each other and the consensus judgments for HSE-2 using $\rho$ and $r_s$. Here, ``Individual" denotes the average $\rho(I_i^p, I_i^q)$ or $r_s(I_i^p, I_i^q)$ of individual estimates between each other, and ``Group" denotes the average $\rho(I_i^p, G_i)$ or $r_s(I_i^p, G_i)$ between each individual estimates and the consensus agreements. These results vary a lot across different groups and are not much better than the machine learning results. Similar to HSE-1, this suggests the potential usefulness of using a machine learning model to help with story point estimation.

\begin{table*}[!tbh]
    \centering
    \caption{Direct judgment section response evaluation for each participant (HSE-2) using using Pearson's Correlation Coefficient ($\rho$) and Spearman's Rank Correlation Coefficient ($r_s$) scores, where ``Individual" denotes the inter-rater reliability: the average $\rho(I_i^p, I_i^q)$ or $r_s(I_i^p, I_i^q)$ of individual estimates between each other, and ``Group" denotes reliability against ground truth: the average $\rho(I_i^p, G_i)$ or $r_s(I_i^p, G_i)$ between each individual estimates and the consensus agreements.}
    \begin{tabular}{|c|c|c|c|}
    \hline
        Group & Type & \makecell{Average\\ $\rho$} & \makecell{Average\\ $r_s$}\\ \hline
        {\multirow{2}{*}{1}} & Individual & 0.32 & 0.27\\
        & Group & 0.16 & 0.19\\ \hline
        {\multirow{2}{*}{2}} & Individual & 0.83 & 0.65\\
        & Group & 0.68 & 0.33\\ \hline
        {\multirow{2}{*}{3}} & Individual & 0.48 & 0.59\\
        & Group & 0.48 & 0.58\\ \hline
        {\multirow{2}{*}{4}} & Individual & 0.60 & 0.60\\
        & Group & 0.80 & 0.80\\ \hline
        {\multirow{2}{*}{5}} & Individual & 0.10 & 0.01\\
        & Group & 0.14 & 0.23\\ \hline
        {\multirow{2}{*}{6}} & Individual & -0.26 & -0.17\\
        & Group & 0.10 & -0.07\\ \hline
        % {\multirow{2}{*}{Average}} & Individual & 0.35 & 0.33\\
        % & Group & 0.40 & 0.34\\
    % \hline
    \end{tabular}
    \label{table:direct_eval_HSE2}
\end{table*}

As discussed in Section~\ref{sect:eval}, the (1) inter-rater reliability: the $p_o$ and $\kappa$ between each two raters (e.g. $p_o (I_{ij}^p, I_{ij}^q),\,\forall p\neq q$); and (2) the reliability against ground truth: the $p_o$ and $\kappa$ between each rater and a set of ground truth story points (e.g. $p_o (I_{ij}^p, G_{ij}),\,\forall p$) were calculated for the comparative judgments in  Table-\ref{table:RQ6_HSE2}. Meanwhile, $p_o$ and $\kappa$ for the direct estimates were derived using $\hat{I}_{ij}^p = \text{sgn}(I_i^p-I_j^p)$ serving as the comparative judgments. In Table-\ref{table:RQ6_HSE2}, the rows marked ``Individual" show the inter-rater reliability: the average of $p_o$ and $\kappa$ for each participant compared to each others' individual judgments. The rows marked ``Group" show the reliability against ground truth: the average of $p_o$ and $\kappa$ for each participant compared against their final group judgments. 

While the evaluation metrics show some variance across groups, groups show comparable agreement and consistency in their comparative judgments and direct judgments. Most notably, participants show similar performances between inter-rater reliability and the reliability against ground truth. But the comparative judgments section sees a notable improvement in the reliability against ground truth when compared to their inter-rater reliability. 

\begin{table*}[!t]
    % \tiny
    \centering
    \renewcommand{\arraystretch}{1.25}
    \caption{Comparing the reliability between direct and comparative judgments (RQ6, HSE-2) for each participant using their relative agreement ($p_o$) and $kappa$ scores for inter-rater reliability and the reliability against ground truth.}
    % \hspace*{-1cm}
    \begin{tabular}{|c|c||c|c||c|c|}
    \hline
        {\multirow{3}{*}{Group}} & {\multirow{3}{*}{Type}} & \multicolumn{4}{c|}{Average metrics}\\ \cline{3-6}
        &  & \multicolumn{2}{c|}{Direct} & \multicolumn{2}{c|}{Comparative}\\ \cline{3-6}
         &  & $p_o$  &  $\kappa$  &  $p_o$  &  $\kappa$ \\ \hline
        {\multirow{2}{*}{1}} & Individual & 0.68 & 0.19 & 0.54 & 0.07\\
        & Group & 0.64 & 0.33 & 0.48 & -0.03\\ \hline
        {\multirow{2}{*}{2}} & Individual & 0.93 & 0.85 & 0.50 & 0.00\\
        & Group & 0.68 & 0.28 & 0.65 & 0.28\\ \hline
        {\multirow{2}{*}{3}} & Individual & 0.81 & 0.62 & 0.47 & -0.01\\
        & Group & 0.84 & 0.66 & 0.63 & 0.25\\ \hline
        {\multirow{2}{*}{4}} & Individual & 0.86 & 0.71 & 0.89 & 0.73\\
        & Group & 0.93 & 0.85 & 0.94 & 0.86\\ \hline
        {\multirow{2}{*}{5}} & Individual & 0.50 & 0.00 & 0.30 & -0.30\\
        & Group & 0.61 & 0.22 & 0.55 & 0.20\\ \hline
        {\multirow{2}{*}{6}} & Individual & 0.39 & -0.28 & 0.53 & 0.08\\
        & Group & 0.45 & -0.16 & 0.67 & 0.33\\ \hline
        {\multirow{2}{*}{Average}} & Individual & 0.70 & 0.35 & 0.54 & 0.09\\
        & Group & 0.69 & 0.36 & 0.65 & 0.32\\
        \hline
    \end{tabular}
    \label{table:RQ6_HSE2}
\end{table*}

Overall, participants in HSE-1 showed more consistency in their comparative judgments than their direct judgments. Individual and group judgments made in HSE-2 showed variable consistency across the groups, but showed comparable consistency on average.\\

\rules{ \noindent\textbf{Answer to RQ6: }The participants provided consistent judgments for both sections, generally agreeing with each other and the ground truths. The comparative judgments showed slightly better consistency than the synthetic judgments generated from the direct judgments in HSE-1 but slightly worse consistency than the synthetic judgments generated from the direct judgments in HSE-2. These findings suggest that in story point estimation, comparative judgments are similarly reliable as direct estimates.\\  }

As discussed in Section-\ref{sec:HSE2_study_design}, we also asked the participants some additional questions at the end of HSE-2. The answers to Q1 and Q2 are shown on Table-\ref{table:RQ5_HSE2}. Q3 asked participants if they felt that discussing the comparative labels together before discussing the direct labels had any positive impact on their direct rating section afterwards. 19 our of the 26 participants answered ``Yes" to this question. Finally, Q4 asked if participants felt more confident in their group labels or individual labels. 17 participants answered feeling more confident in their group labels, 1 participant felt more confident in their individual labels, and 8 participants felt equally confident about both sections.

%%%%%%%%%%%%%%%%%%%%%%%%%%%%%%%%%%%%%%%%%%%%%%%%%%%%%%%%%%%%%%%%%%%%%%%%%%%%%%%%%%%
\section{Threats to validity}
\label{sec:threats}

\begin{itemize}

    \item \textbf{Construct validity: }The proposed models' performances are compared against other baseline works. These works were reproduced using either the instructions described in those works or the code made publicly available. Although the reproduced results resemble the reported results in terms of MAE scores, there could be differences in the hyperparameters not mentioned that could result in different $\rho$ or $r_s$ scores.

    % \item \textbf{Conclusion validity: }No threats to the work's conclusion validity were found.

    % \item \textbf{Internal validity: }No threats to the work's Internal validity were found. \yiming{you need to write something here. This statement is not accepted by reviewers}

    \item \textbf{Internal validity: }Although the human subject experiments show less annotation time for comparative judgments when compared to direct estimates, the shorter annotation time can be due to different reasons including less cognitive burden and that comparative judgments are easier to make. There is a internal validity threat that other factors could also contribute to the difference in annotation time.

    \item \textbf{External validity: }The unique contexts of Agile story point estimation for each project, and the varying experiences of human developers could lead to a variety of different outcomes when a comparative approach is taken in a different practical setting, especially since these developers are used to following the traditional methods. As such, more in-depth human experiments recording the human efforts, and the generalizability and usability of such results are required to verify and validate this work's findings beyond any reasonable doubt.
    
\end{itemize}

\section{Conclusion}
\label{sec:future_work_and_conclusion}
The goal of this work is to \goalStatement.
To that end, we propose a framework to model comparative judgments instead of direct story points. We used existing story point estimation data and models to simulate the process and to demonstrate the potential benefits of our approach. Results show that our approach performed similarly to, if not better than, its regression counterpart and other state-of-the-art baseline models. We also conduct cross-project experiments to show that although our proposed models outperform state-of-the-art models for cross-project settings as well, the overall performances decrease significantly, and as such, project-specific training data is still required for higher performance from machine learning models. Finally, we conduct human-subject experiments to observe that human judges do provide comparably accurate yet quicker and more confident judgments through a comparative approach.
Our results from both the machine learning and human subject experiments indicate a promising reduction in human annotation effort when adopting comparative judgments for story point estimation.

\section{Data availability}
The code and link to the data for our approach here are available at \url{https://github.com/hil-se/EfficientSPEComparativeLearning}. This repository includes links to raw text and story point data without any preprocessing.

\section*{Acknowledgment}
This work was funded by NSF grant 2245796.

\newpage

\bibliographystyle{spmpsci}      % mathematics and physical sciences
\bibliography{Bibliography}   % name your BibTeX data base

\clearpage
\appendix

\end{document}